%% file: main.tex
\documentclass{article}

\usepackage[preprint]{neurips_2026}

\usepackage[utf8]{inputenc} 
\usepackage[T1]{fontenc}    
\usepackage{hyperref}       
\usepackage{amsmath}
\usepackage{bm} 
\usepackage{url}            
\usepackage{booktabs}       
\usepackage{amsfonts}       
\usepackage{nicefrac}       
\usepackage{microtype}      
\usepackage{xcolor}         
\usepackage{graphicx}
\usepackage{color}
\usepackage{transparent}
\usepackage{multirow}
\usepackage{graphicx}
\usepackage{booktabs}  
\usepackage{algorithm}
\usepackage{algpseudocode}
\usepackage[table]{xcolor}
\newcommand{\best}[1]{\colorbox{red!35}{\strut #1}}
\newcommand{\second}[1]{\colorbox{red!15}{\strut #1}}
\usepackage{soul}
\title{3D Skew-Normal Splatting}

\author{%
  Xiangru Wu \\
  Fudan University \\
  \texttt{xiangruwu22@m.fudan.edu.cn} \\
  \And
  Ke Fan \\
  Fudan University \\
  \texttt{kfan21@m.fudan.edu.cn} \\
  \And
  Yanwei Fu \\
  Fudan University \\
  \texttt{yanweifu@fudan.edu.cn} \\
}

\begin{document}
\maketitle
\begin{abstract}
  3D Gaussian Splatting (3DGS) has emerged as a leading representation for real-time novel view synthesis and has been widely adopted in various downstream applications. The core strength of 3DGS lies in its efficient kernel-based scene representation, where Gaussian primitives provide favorable mathematical and computational properties. However, under a finite primitive budget, the symmetric shape of each primitive directly affects representation compactness, especially near asymmetric structures such as object boundaries and one-sided surfaces. Recent works have explored more complex kernel distributions; however, they either remain within the elliptical family or rely on hard truncation, which limits continuous shape control and introduces distributional discontinuities. In this paper, we propose Skew-Normal Splatting (SNS), which adopts the Azzalini Skew-Normal distribution as the fundamental primitive. By introducing a learnable and bounded skewness parameter, SNS can continuously interpolate between symmetric Gaussians and Half-Gaussian-like shapes, enabling flexible modeling of both sharp boundaries and interior regions. Moreover, SNS preserves analytical tractability under affine transformations and marginalization. This property allows seamless integration into existing Gaussian Splatting rasterization pipelines. Furthermore, to address the strong coupling between scale, rotation, and skewness parameters, we introduce a decoupled parameterization and a block-wise optimization strategy to enhance training stability and accuracy. Extensive experiments on standard novel-view synthesis benchmarks show that SNS consistently improves reconstruction quality over Gaussian and recent non-Gaussian kernels, with clearer benefits on sharp boundaries and thin or one-sided structures. 
\end{abstract}

\section{Introduction}
In the field of novel view synthesis, the increasing demand for real-time interactivity has exposed the limitations of traditional NeRF-based~\cite{DBLP:conf/eccv/MildenhallSTBRN20,DBLP:journals/cacm/MildenhallSTBRN22,DBLP:conf/iccv/BarronMTHMS21,DBLP:conf/cvpr/BarronMVSH22} methods regarding rendering efficiency and training overhead. Since its introduction, 3D Gaussian Splatting~\cite{DBLP:journals/tog/KerblKLD23} has rapidly emerged as a dominant alternative to early NeRF approaches. By leveraging explicit scene representations and fast, differentiable rendering, it achieves a superior balance between visual fidelity and computational performance and has been widely used in fields such as robotics~\cite{DBLP:conf/cvpr/KeethaKJYSRL24}, autonomous driving~\cite{DBLP:conf/cvpr/ZhouLSWS024}, and cultural heritage preservation~\cite{DBLP:journals/corr/abs-2409-19039}. The success of 3D Gaussian Splatting is attributed to its use of Gaussian functions as kernels, which provide powerful approximation capabilities alongside mathematical and computational efficiency~\cite{zwicker2001ewa, zwicker2002ewa}.
Nevertheless, the universal approximation property of kernel-based methods remains largely a theoretical guarantee, typically predicated on the availability of an infinite and sufficiently dense distribution of kernels. In the practical regime of a finite primitive budget, the distributional form of each kernel directly dictates representation compactness and reconstruction accuracy. Symmetric Gaussian kernels often require multiple overlapping primitives to approximate sharp boundaries, leading to redundant representations and blurred reconstructions.

Recent efforts~\cite{DBLP:conf/cvpr/ZhuYH025, DBLP:conf/cvpr/Li0SC25, DBLP:conf/cvpr/HamdiMMQLVGV24} have investigated the use of more sophisticated distributions as alternatives to the standard Gaussian. Despite some success, these methods still have limitations. Both Generalized Exponential Splatting (GES)~\cite{DBLP:conf/cvpr/HamdiMMQLVGV24} and Student Splatting and Scooping (SSS)~\cite{DBLP:conf/cvpr/ZhuYH025} propose to extend the Gaussian primitive by modifying the radial density post-whitening; however, such methods are still restricted to the elliptical distribution family, which imposes a persistent radial symmetry on the representation. Among these efforts, HGS~\cite{DBLP:conf/cvpr/Li0SC25} adopts the Half-Gaussian distribution, representing an early attempt to introduce asymmetric kernels into scene modeling. However, this approach relies on a hard truncation scheme that simply divides a Gaussian distribution at its centroid, thereby offering only limited expressivity in terms of skewness. Furthermore, such truncation inherently introduces discontinuities into the primitive’s distribution.

To further characterize the asymmetry that is prevalent in real data, many works in Statistics~\cite{Kundu2014GeometricSN,ElalOlivero2010ALPHASKEWNORMALD,Mameli2011AGO,Mudholkar2000TheED,ArellanoValle2006OnTU,arnold2000hidden} extend the Gaussian distribution with additional shape parameters to capture skewness. These distributions are widely used in fields such as economics and finance, biomedicine, and spatial statistics~\cite{vernic2006multivariate,carmichael2013asset,fruhwirth2010bayesian,rimstad2014skew,zhang2010spatial}. Among these works, Azzalini's Skew-Normal distribution~\cite{Azzalini_1999} is widely regarded as a classical extension of the Gaussian distribution, due to its simplicity and clear probabilistic construction, providing a useful reference for our work.

\begin{figure}[tbp]
    \centering
    \includegraphics[width=\textwidth]{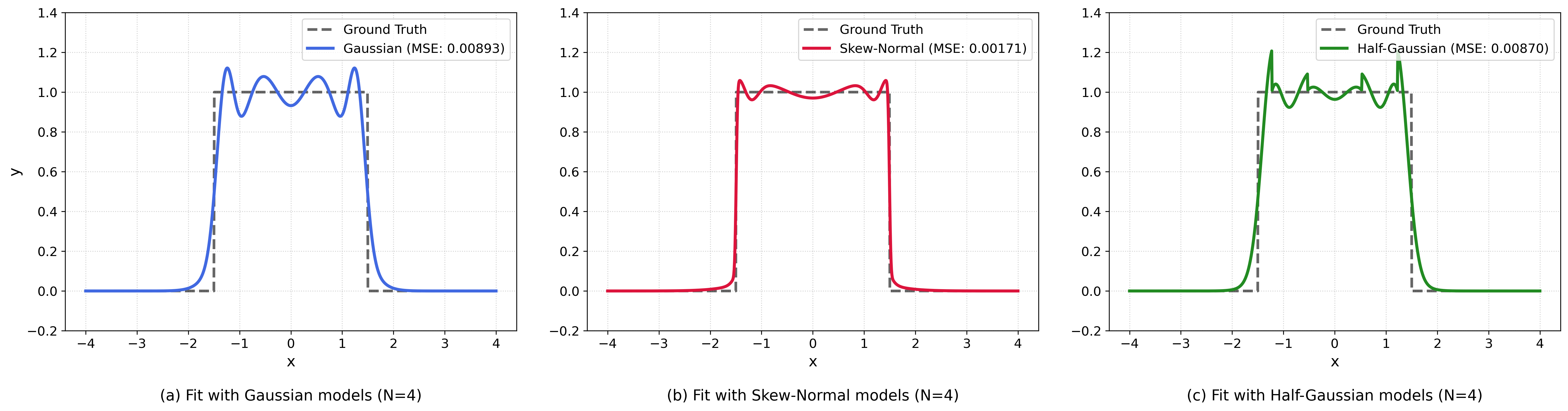}
    \vspace{-0.1in}
    \caption{\textbf{Reconstruction results.} We fit a square wave with different basis models  using a sum of four primitives from: \textbf{(a)} Gaussian, \textbf{(b)} Skew-Normal, and \textbf{(c)} Half-Gaussian distributions. Their parameters are optimized via gradient descent. Our Skew-Normal model achieves the lowest Mean-Squared Error (MSE), demonstrating its better capability in approximating signals with sharp discontinuities.  \label{fig:1d_interpolation} }
    \vspace{-0.25in}
\end{figure}

To this end, we propose Skew-Normal Splatting (SNS), which employs the Skew-Normal distribution of Azzalini as the underlying primitive for scene modeling. By subsuming the standard Gaussian and half-Gaussian distributions as special or limiting cases, the Skew-Normal distribution provides a continuous and tunable mechanism for asymmetric modeling.
On one hand, it offers the flexibility to  capture non-symmetric structures such as sharp edges and corners; on the other, it maintains the capacity to effectively represent the interior regions of objects. As Fig. \ref{fig:1d_interpolation} shows, we fit the same 1D square wave using Gaussian, Skew-Normal, and Half-Gaussian kernels, respectively. For a fair comparison, all methods are restricted to 4 components and optimized via gradient descent. The Gaussian model exhibits boundary ringing-like artifacts, including overshoot and undershoot near step transitions. The Half-Gaussian method offers a marginal improvement but introduces discontinuities and sharp peaks at the step changes. In contrast, SNS can effectively fit both sharp edges and flat portions of the square wave, demonstrating its superior representational capacity.

Furthermore, the SNS remains analytically tractable under affine transformations and marginalization. Leveraging this mathematical convenience, we re-derive the projection and integration routines required to incorporate SNS into the splat-based rasterization pipeline, thereby enabling highly efficient computation.

Beyond its representational benefits, the transition to asymmetric primitives introduces new analytical and optimization challenges. We observe that the location parameter of a Skew-Normal distribution does not coincide with its spatial mean. We thus derive a corrected operational center to ensure accurate bounding-box localization. Furthermore, to resolve the optimization instability caused by the coupling between orientation and skewness, we introduce a decoupled parameterization in the primitive's canonical coordinate system, supported by a block-wise alternating update strategy. This design effectively reduces parameter interference, ensuring stable convergence and improved generalization. Formally, our contributions include:

\begin{itemize}
    \item We introduce Azzalini’s Skew-Normal distribution to Gaussian Splatting, which has varying degrees of skewness for 3D scene modeling and can simultaneously handle object boundaries and interiors.
    \item We derive a complete analytical framework for Skew-Normal kernels, including closed-form affine projection, corrected footprint localization, and efficient gradient computation for rasterization, 
    preserving compatibility with existing 3DGS pipelines.
    \item We identify the optimization coupling between geometric and skewness parameters and propose a decoupled reparameterization along with a block-wise alternating update scheme to enhance training stability.
    \item SNS achieves consistent PSNR gains over recent non-Gaussian variants and improves visual preservation of sharp structures and object completeness.
\end{itemize}
   
\begin{figure}[t]
    \centering
    \includegraphics[width=\textwidth]{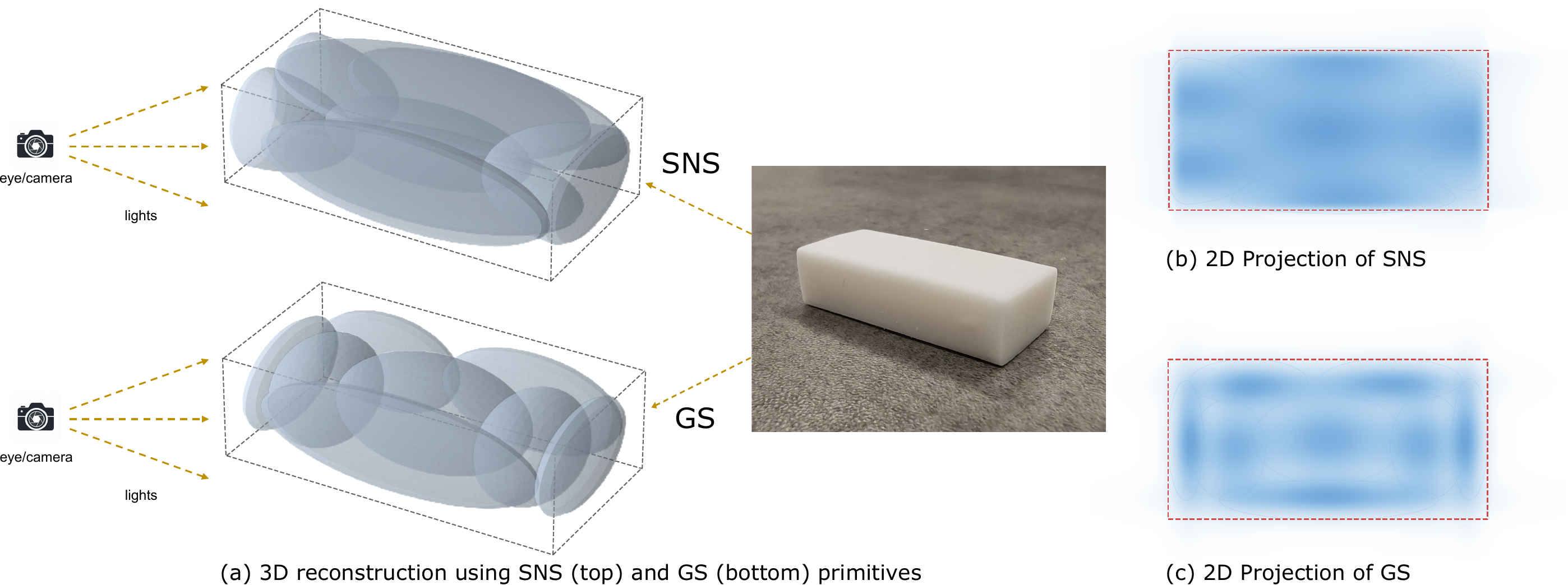}
     \vspace{-0.1in}
    \caption{\textbf{Comparison between Skew-Normal and Gaussian primitives. } Standard Gaussian primitives are symmetric and often require multiple overlapping kernels to approximate sharp or one-sided structures. By contrast, SNS introduces a learnable skewness parameter, enabling diverse asymmetric primitives to accurately fit sharp corners and flat surfaces.}
    \label{fig:teaser}
    \vspace{-0.15in}
\end{figure}

\section{Related Work}
\noindent\textbf{NeRF methods.}
Early approaches to 3D reconstruction such as SfM~\cite{DBLP:conf/cvpr/SchonbergerF16,DBLP:conf/eccv/WeiZLFX20} and MVS~\cite{DBLP:conf/eccv/YaoLLFQ18,DBLP:journals/tmlr/CaoRF22} typically rely on explicit geometric estimation. Recently, Neural Radiance Fields (NeRF)~\cite{DBLP:conf/eccv/MildenhallSTBRN20,DBLP:journals/cacm/MildenhallSTBRN22} implicitly learn scene geometry and appearance by parameterizing volumetric density and color through neural networks, utilizing volume rendering to synthesize novel views. Mip-NeRF~\cite{DBLP:conf/iccv/BarronMTHMS21} and Mip-NeRF 360~\cite{DBLP:conf/cvpr/BarronMVSH22} reduce aliasing and extend NeRF to unbounded scenes, while explicit or hybrid representations such as Plenoxels~\cite{DBLP:conf/cvpr/Fridovich-KeilY22} and K-Planes~\cite{DBLP:conf/cvpr/Fridovich-KeilM23} improve training and rendering efficiency.
Beyond reconstruction, NeRF has been extended to sparse-view synthesis~\cite{DBLP:conf/cvpr/YuYTK21,DBLP:conf/cvpr/WangWGSZBMSF21}, object-centric decomposition~\cite{DBLP:conf/iclr/YuG022,DBLP:conf/nips/QiYZ23}, neural surface reconstruction~\cite{DBLP:conf/nips/YarivGKL21,DBLP:conf/nips/WangLLTKW21}, and dynamic scene modeling~\cite{DBLP:conf/cvpr/PumarolaCPM21,DBLP:journals/tog/ParkSHBBGMS21}, substantially advancing neural 3D representations. Despite their strong rendering quality, NeRF-based methods typically rely on dense ray sampling and implicit network evaluation, which makes training and rendering computationally expensive compared with explicit splatting-based representations.

\noindent\textbf{Splatting methods.} The 3DGS~\cite{DBLP:journals/tog/KerblKLD23} models the scene with optimizable Gaussian elements and projects them onto the image plane through an efficient splatting rasterizer, thereby significantly accelerating the rendering process.
Building upon this paradigm, several works have improved 3DGS across multiple dimensions. Mip-Splatting~\cite{DBLP:conf/cvpr/YuCHS024} focuses on anti-aliasing and multi-scale rendering, while Scaffold-GS~\cite{DBLP:conf/cvpr/0005YXX0L024} and Octree-GS~\cite{10993308} introduce hierarchical structures to enhance representation efficiency. To optimize the growth and pruning strategies of Gaussian primitives, Pixel-GS~\cite{DBLP:conf/eccv/ZhangHLHZ24} and AbsGS~\cite{10.1145/3664647.3681361} refine the density control mechanisms. Furthermore, 3DGS has been extended to a broader range of applications: GaussianEditor~\cite{DBLP:conf/cvpr/WangF0X024,DBLP:conf/cvpr/ChenCZWYWCYLL24} enables scene editing within the Gaussian representation, and 4DGS~\cite{DBLP:conf/cvpr/WuYFX0000W24,DBLP:conf/iclr/YangYP024} further generalizes the framework to dynamic scene modeling. Collectively, these advancements have established 3DGS as a powerful alternative for efficient 3D scene representation and rendering.
Beyond these domains, the framework has also been applied to 3D content generation~\cite{DBLP:conf/iclr/TangRZ0Z24,DBLP:conf/cvpr/YiFWWX000W24,DBLP:conf/eccv/TangCCWZL24, DBLP:conf/eccv/XuSWCYPSW24}, language-driven scene understanding~\cite{DBLP:conf/nips/WuMLWSC0FD0024,DBLP:conf/cvpr/Qin0ZWP24,11442667}, and 3D scene segmentation~\cite{DBLP:conf/aaai/CenF0X00025,DBLP:conf/eccv/YeDYK24}.
More recently, several works have explored extending 3DGS by modifying the underlying kernel functions of the Gaussian primitives. For instance, GES~\cite{DBLP:conf/cvpr/HamdiMMQLVGV24} adopts the Generalized Exponential Function, while SSS employs the Student’s t-distribution~\cite{DBLP:conf/cvpr/ZhuYH025}. Pushing this further, 3D-HGS~\cite{DBLP:conf/cvpr/Li0SC25} introduces the half-Gaussian distribution to facilitate the representation of more irregular primitives.
Unlike GES and SSS, which modify the radial profile after whitening and therefore remain largely symmetric in canonical space, our SNS introduces directional skewness as an intrinsic primitive parameter. Unlike 3D-HGS, SNS avoids hard truncation and provides a continuous path from Gaussian-like to half-Gaussian-like shapes.

\section{Method}
\subsection{Gaussian Splatting}
\noindent\textbf{Scene Representation.}
3DGS models the scene using 3D Gaussian kernels. Each kernel is parameterized by a center position $\mu \in \mathbb{R}^{3}$, a 3D covariance matrix $\Sigma \in \mathbb{R}^{3 \times 3}$, an opacity value $o \in [0, 1]$, and view-dependent color features $c$. To ensure positive semi-definiteness during optimization, $\Sigma$ is factorized into a scaling matrix $S \in \mathbb{R}^{3 \times 3}$ and a rotation matrix $R\in \mathbb{R}^{3 \times 3}$, formulated as: $\Sigma = R S S^\top R^\top$. These 3D Gaussians are projected onto the 2D screen space to render images.  Given a viewing transformation matrix $W$ and the Jacobian of the affine projection approximation $J$, the 2D covariance matrix, denoted by $\Sigma'$, can be computed as: $\Sigma' = J W \Sigma W^\top J^\top$.

\noindent\textbf{Volumetric Alpha-Blending.}
The rendering process evaluates the final color $C(u)$ at pixel $u$ through a tile-based rasterizer. Analogous to volumetric rendering, 3DGS performs alpha-blending by sorting the projected Gaussians along the ray corresponding to the pixel from front to back. For a set of $\mathcal{N}$ ordered, intersecting Gaussians, the pixel color is aggregated as~\cite{DBLP:journals/tog/KerblKLD23}:
\begin{equation}
    C(u) = \sum_{i=1}^{\mathcal{N}} c_i a_i(u) \prod_{j=1}^{i-1} \left(1 - a_j(u)\right),
    \label{eq:rendering}
\end{equation}
where $c_i$ is the color evaluated from the SH coefficients. The term $a_i(u)$ denotes the spatial influence of the $i$-th Gaussian evaluated at pixel $u$, which is obtained by multiplying the learned opacity $o_i$ with the 2D Gaussian kernel:
\begin{equation}
    a_i(u) = o_i \exp \left( -\frac{1}{2}(u - \hat{\mu}_i)^\top \hat{\Sigma}_i^{-1} (u - \hat{\mu}_i) \right),
    \label{eq:influence}
\end{equation}
where $\hat{\mu}_i$ is the projected 2D center position.

\subsection{3D Skew-Normal Kernel}
\label{sec:sn-kernel}
\begin{figure}[tbp]
    \centering
    \includegraphics[width=\textwidth]{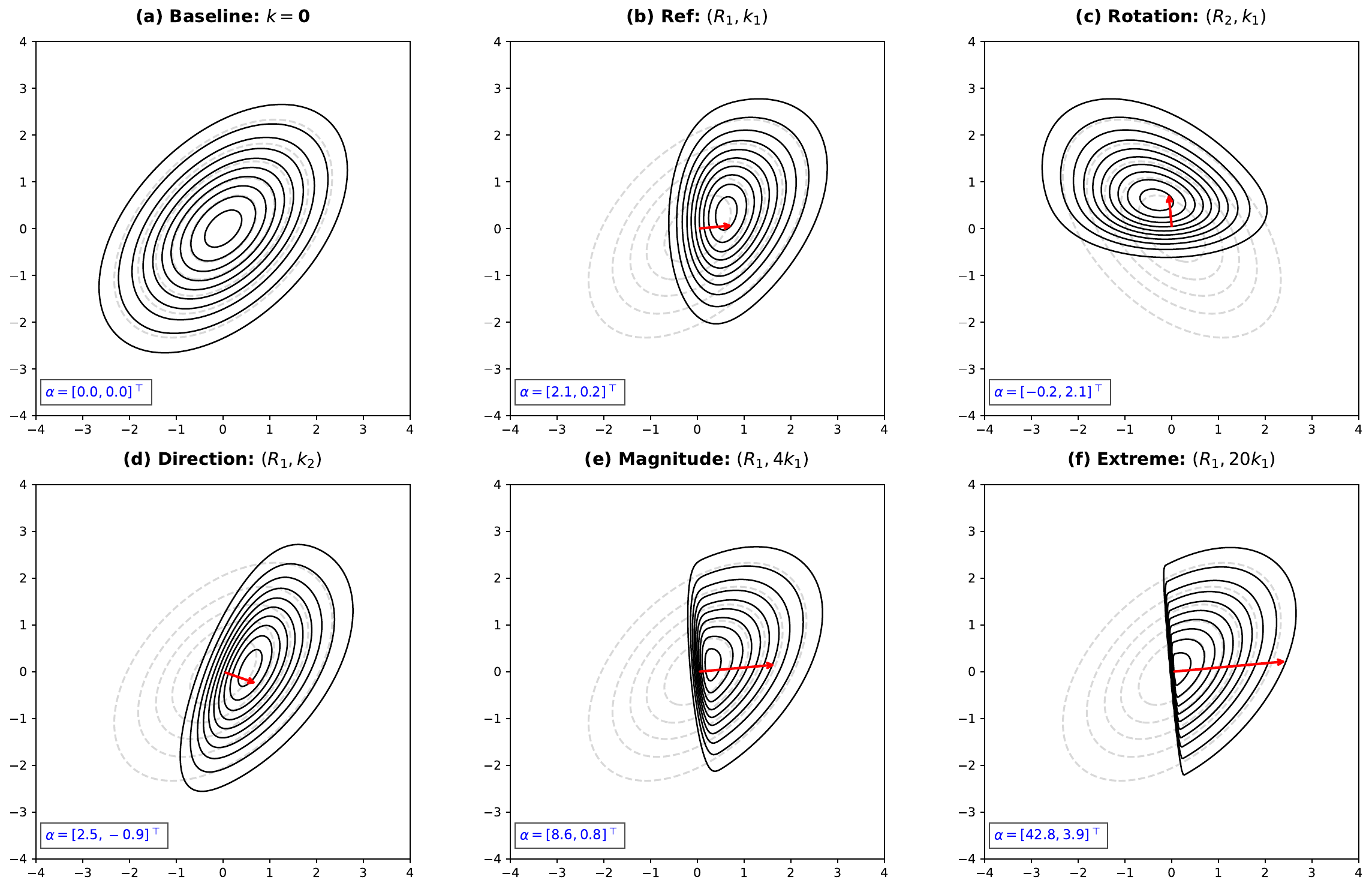}
    \vspace{-0.1in}
    \caption{\textbf{Parameter decoupling of bivariate SND.} 
    These panels demonstrate the continuous shape interpolation and decoupled control of the Skew-Normal density. \textbf{(a)} Baseline standard symmetric Gaussian. \textbf{(b vs. d)} Directional control via the orientation of $k$. \textbf{(b, e, f)} Magnitude variation and rapid convergence. \textbf{(b vs. c)} Intrinsic nature demonstrated via rigid global rotation $R$. In each panel, the dashed line indicates the symmetric Gaussian base and the solid lines depict the Skew-Normal density. All panels share an identical scaling matrix $S$.     
    \label{fig:shape_interpolation} }
    \vspace{-0.1in}
\end{figure}

We inherit Azzalini's definition of the Skew-Normal Distribution (SND), which provides directional asymmetry through a new parameter $\alpha$.
A d-dimensional SND random variable $Y$ is denoted as $Y \sim \mathcal{SN}_{d}(\mu, \Omega, \alpha)$, where the parameters are called location, scale matrix, and slant~\cite{Azzalini_1999}, respectively. The slant parameter $\alpha$ is highly coupled with $\Omega$. When $\alpha$ is kept fixed, applying a spatial rotation to the model does not only reorient the primitive but also changes its intrinsic geometric shape.
To decouple the shape from the orientation, we propose an unnormalized and reparameterized 3D Skew-Normal kernel, which is defined by a location vector $\mu \in \mathbb{R}^3$, a scale matrix $\Omega = RSS^\top R^\top \in \mathbb{R}^{3 \times 3}$, and a learnable latent variable $k \in \mathbb{R}^3$:
\begin{equation}
    SN(x; \mu, \Omega, k) = G_3(x - \mu; \Omega) \, \Phi \left( \alpha^\top \omega^{-1} (x - \mu) \right),
    \label{eq:sg_3d}
\end{equation}
where $\alpha = \omega R S^{-1} k$, $\omega = \left( \Omega \odot \mathbf{I}_3 \right)^{1/2}$, $G_3(x - \mu; \Omega) = \exp \left( -\frac{1}{2} (x - \mu)^\top \Omega^{-1} (x - \mu) \right)$, and $\Phi(\cdot)$ denotes the cumulative distribution function (CDF) of the standard univariate normal distribution. The symbol $\odot$ refers to the Hadamard product. This formulation provides four main advantages:

\noindent\textbf{Continuous Shape Interpolation.}
The Skew-Normal distribution includes the Gaussian and half-Gaussian as limiting cases, providing a natural continuum between symmetric and maximally skewed geometries. 
It reduces to a Gaussian ellipsoid when $\alpha=\mathbf{0}$, and approaches a half-ellipsoid as $\|\alpha\|\to\infty$. As illustrated in Fig. \ref{fig:shape_interpolation}, this continuous interpolation enables a single kernel to represent a broad spectrum of geometric structures. By increasing the magnitude from $\|k_1\| \approx 2.24$ to $4\|k_1\| \approx 8.94$, the kernel reshapes into a sharp half-ellipsoid. Crucially, further scaling to $20\|k_1\|$ yields a slight visual change, demonstrating that the skewness converges rapidly to its limiting form. 
Such flexibility extends the expressive power of Gaussian kernels while preserving a compact and principled parameterization.

\noindent\textbf{Bounded Skewness.}
The shape of the SND is influenced by the slant $\alpha$ and the scale matrix $\Omega$. To provide a single, scalar measure of skewness, Mardia's measure of skewness~\cite{Azzalini_1999} is expressed as:
\begin{equation}
\label{eq:Mardia_skewness}
    \gamma_{1,d}^M = \left( \frac{4-\pi}{2}\right)^2  \left( \frac{(2/ \pi) \alpha_*^2}{1+ (1-2/ \pi) \alpha_*^2}\right)^3,
\end{equation}
where $\alpha_* = \left(\alpha^\top \bar{\Omega} \alpha\right)^\frac{1}{2} \in \left[0,\infty \right)$.
As shown in Eq.(\ref{eq:Mardia_skewness}), $\gamma_{1,d}^M$ is monotonically increasing with respect to $\alpha_*$. When $\alpha_* = 0$, $\gamma_{1,d}^M = 0$; when $ \alpha_* \to \infty$, $\gamma_{1,d} \to \frac{2(\pi -4)^2}{(\pi - 2)^3}$, which is approximately $0.99067$. Therefore, the skewness of the SND is confined to a bounded region, preventing the distribution from being excessively stretched. This bounded property is important for the 3DGS framework. 3DGS relies on tile-based rasterization, which requires primitives to have a spatially compact footprint for efficient bounding box calculation. An unbounded asymmetric tail would cause a primitive to intersect with an excessive number of tiles, degrading rendering performance. On the other hand, the local affine approximation (via Jacobian $J$) employed in 3DGS is fundamentally predicated on a localized spatial footprint. An arbitrarily extended tail would violate this assumption and lead to instability during optimization.

\noindent\textbf{Decoupled Skewness Measure.} By defining the slant parameter as $\alpha = \omega R S^{-1} k$, we decouple the asymmetry of the distribution from its affine transformation components (i.e., $R$ and $S$). 
Under this reparameterization, the term $\alpha_*$ in Eq.(\ref{eq:Mardia_skewness}) reduces to the $L_2$-norm of the skew vector $k$, implying that $\gamma_{1,d}^M$ is solely determined by $\|k\|_2$. It reveals that the magnitude of $k$ comprehensively encapsulates the departure from normality. In Fig. \ref{fig:shape_interpolation}, panels (b) and (c) share the same $k_1$ but differ in rotation $R$. The resulting density in (c) is a rigid rotation of (b), verifying that $k$ is an intrinsic property defined in the primitive's canonical coordinate system; varying $R$ only alters global orientation without deforming the asymmetric kernel itself. Additionally, shifting $k$'s orientation (e.g., rotating $k_1 = (2, -1)^\top$ to $k_2$ while maintaining $\|k_1\|=\|k_2\|$)  dictates the specific axis of the skew, remapping the kernel's asymmetry to a new direction.

\noindent\textbf{Closed-form 2D Projection.}
The SND is closed under affine transformations. We derive the analytical form by projecting a 3D primitive to the 2D image plane. This derivation requires an affine transformation matrix $A = JW \in \mathbb{R}^{2 \times 3}$. Formally, a 2D Skew-Normal kernel is described by:
\begin{equation}
\label{eq:2D_krenl}
    SN^{2D}(u;\mu^{2D}, \Omega^{2D}, m^{2D}) = G_2(u - \mu^{2D}; \Omega^{2D}) \, \Phi \left( (m^{2D})^\top (u - \mu^{2D}) \right).
\end{equation}
where $u \in \mathbb{R}^2$ denotes the pixel coordinate, 

\begin{equation}
\label{eq:affine_transform_params}
\begin{aligned}
    \mu^{2D} &= A \mu,\quad
    \Omega^{2D} = A \Omega A^\top, \\
    m^{2D} &= \frac{(\Omega^{2D})^{-1} A RSk}{\sqrt{1 + k^\top k - (A RSk)^\top (\Omega^{2D})^{-1} A RSk}}.
\end{aligned}
\end{equation}
The reparameterization of $\alpha$ is essential; otherwise, the diagonal matrix
$\omega$ remains coupled with the inverse 2D covariance, which complicates gradient backpropagation. Although the reparameterized expression for $m^{2D}$ appears computationally heavy, we reduce the training cost by restructuring the backward propagation paths through algebraic reductions. Details are in appendix ~\ref{sec:closure} and ~\ref{sec:backward_pass}.

\subsection{Forward Modeling of Splatting}
Restricting primitives strictly to a positive density space bounds their representational capacity. Following SSS, we allow signed opacity values $o_i \in [-1, 1]$, enabling primitives to both contribute color and subtract volumetric density from the scene. This subtractive capability can represent complex topologies with fewer kernels. At a pixel $u$, the front-to-back alpha blending evaluates as:
\begin{equation}
    C(u) = \sum_{i \in \mathcal{N}} c_i o_i SN^{2D}_i(u) \prod_{j=1}^{i-1} \left(1 - o_j SN^{2D}_j(u)\right),
\end{equation}
where $\mathcal{N}$ is the depth-sorted set of intersecting kernels.

Traditional splatting frameworks, such as 3DGS and SSS, anchor their tile-based bounding boxes to the nominal location parameter $\mu^{2D}$, which is the mean of the distribution. As shown in Fig. \ref{fig:shape_interpolation}, $\mu^{2D}$ deviates from the spatial mean because the slant parameter $\alpha$ shifts the probability mass. Relying on $\mu^{2D}$ for bounding boxes localization is ineffective in capturing the spatial density of skewed objects. To address this, we physically shift the 2D footprint to the true spatial mean, deriving the corrected operational center, $\mu_{bbox}$, as a deterministic spatial offset:
\begin{equation}
    \mu_{bbox} = \mu^{2D} + \sqrt{\frac{2}{\pi}} \cdot \frac{ARS k}{\sqrt{1 + k^\top k}}.
\end{equation}

\subsection{Stable Optimization via Parameter Decoupling}
Optimizing Skew-Normal kernels is particularly challenging due to the coupling between the geometric and skewness parameters. We update the positional parameters $\mu$ using Stochastic Gradient Hamiltonian Monte Carlo (SGHMC)~\cite{DBLP:conf/cvpr/ZhuYH025}, which introduces stochasticity to help the model escape local minima. The greater challenge, however, comes from the slant parameter $\alpha$, as it deeply couples the rotation $R$, scaling $S$, and the latent variable $k$.
We decouple them from two perspectives:

\noindent\textbf{Internal parameter dependencies.}
As in Sec.~\ref{sec:sn-kernel}, the magnitude of parameter $k$ explicitly determines the skewness. We therefore  factorize $k$  into an independent magnitude scalar $m_k \in \mathbb{R}_+$ and a direction vector $d_k\in \mathbb{R}^3(\| d_k \|_2 = 1)$. 
We further parametrize the magnitude and direction as:
\begin{equation}
    m_k = \frac{8}{1+exp(-\frac{x}{6})}, \quad d_k = \frac{v}{\|v \|_2+\epsilon},
\end{equation}
where $x \in \mathbb{R}, v\in \mathbb{R}^3$.
This structural decomposition isolates the skew direction and skew intensity, allowing the model to optimize them independently.

\noindent\textbf{Inter-parameter interference.}
Although decomposing $k$ into magnitude and direction improves stability, updating shape parameters $(R, S, m_k, d_k)$ simultaneously still leads to suboptimal convergence. Specifically, adjustments to the base ellipsoid's orientation frequently conflict with the skew direction. To address this, we divide the training process into two phases. First, all parameters ($R, S, m_k, d_k$) are jointly optimized using standard Adam to establish a reasonable global structure. In the second phase, a Block Coordinate Descent (BCD) strategy is activated. The optimization follows a cyclic alternating schedule: in the first part of each cycle, the ellipsoidal base parameters ($R, S$) are updated while the skew parameters ($m_k, d_k$) are frozen; in the second part, ($m_k, d_k$) are refined while ($R, S$) are held constant.

While SGHMC is utilized for the positional parameters $\mu$, the Adam optimizer is employed for all remaining variables. SGHMC naturally provides stochastic global search, preventing positional parameters from collapsing into local minima. Meanwhile, the alternating update scheme governs the shape parameters, ensuring geometric stability throughout training.

\section{Experiments}
\begin{table}[t]
\caption{
    \textbf{Quantitative comparison with baselines.} The best and second-best results are highlighted in dark red and light red, respectively.\label{tab:main}
}
\setlength{\fboxsep}{1pt} 
\renewcommand{\arraystretch}{1.2}
\resizebox{\linewidth}{!}{
\begin{tabular}{l|ccc|ccc|ccc}
\toprule
    Dataset & \multicolumn{3}{c|}{Mip-NeRF360}  & \multicolumn{3}{c|}{Tanks\&Temples} & \multicolumn{3}{c}{Deep Blending}\\
    Method & PSNR $\uparrow$   & SSIM$\uparrow$   & LPIPS$\downarrow$ & PSNR $\uparrow$   & SSIM$\uparrow$   & LPIPS$\downarrow$ & PSNR $\uparrow$   & SSIM$\uparrow$   & LPIPS$\downarrow$ \\
    \midrule 
    Mip-NeRF & 29.23 & 0.844 & 0.207 & 22.22 & 0.759 & 0.257 & 29.40 & 0.901 & 0.245\\
    3DGS & 28.69 & 0.870 & 0.182 & 23.14 & 0.841 & 0.183 & 29.41 & 0.903 & 0.243\\
    GES & 28.72 & 0.794 & 0.250 & 23.34 & 0.836 & 0.198 & 29.82 & 0.901 & 0.252\\
    3DHGS & 29.56 & 0.873 & 0.178 & 24.50 & 0.857 & 0.169 & 29.76 & 0.905 & \second{0.242}\\
    Fre-GS & 27.85 & 0.826 & 0.209 & 23.96 & 0.841 & 0.183 & 29.93 & 0.904 & \best{0.240}\\
    Scaffold-GS & 28.84 & 0.848 & 0.220 & 23.96 & 0.853 & 0.177 & \second{30.21} & 0.906 & 0.254\\
    3DGS-MCMC & 29.89 & \best{0.900} & 0.190 & 24.29 & 0.860 & 0.190 & 29.67 & 0.890 & 0.320\\
    SSS & \second{29.90} & 0.893 & \second{0.145} & \second{24.87} & \second{0.873} & \second{0.138} & 30.07 & \second{0.907} & 0.247\\
    \midrule 
    Ours & \best{30.17 }& \second{0.895} & \best{0.144} & \best{25.08} & \best{0.877} & \best{0.132} & \best{30.30} & \best{0.912} & {0.244 } \\
    \bottomrule
\end{tabular}
}
\end{table}

\noindent\textbf{Datasets and Metrics.} 
We evaluate SNS on 11 scenes across 3 datasets to validate our contributions: 7 scenes from Mip-NeRF 360~\cite{DBLP:conf/cvpr/BarronMVSH22}, 2 scenes from Tanks \& Temples~\cite{DBLP:journals/tog/KnapitschPZK17}, and 2 scenes from Deep Blending~\cite{DBLP:journals/tog/HedmanPPFDB18}. Following the standard evaluation metrics, we employ three metrics: Peak Signal-to-Noise Ratio (PSNR), Structural Similarity Index Measure (SSIM), and Learned Perceptual Image Patch Similarity (LPIPS). We report dataset-level averages, with per-scene results in the appendix.

 \noindent\textbf{Baselines.} (1) kernel-level variants such as GES~\cite{DBLP:conf/cvpr/HamdiMMQLVGV24}, 3D-HGS~\cite{DBLP:conf/cvpr/Li0SC25} and SSS~\cite{DBLP:conf/cvpr/ZhuYH025}, which introduce non-Gaussian primitives; (2) optimization-focused methods like Scaffold-GS~\cite{DBLP:conf/cvpr/0005YXX0L024}, Fre-GS~\cite{DBLP:conf/cvpr/ZhangZXLX24} and 3DGS-MCMC~\cite{DBLP:conf/nips/KheradmandRSSTI24}; and (3) Neural Radiance Field method such as Mip-NeRF~\cite{DBLP:conf/iccv/BarronMTHMS21}. For a fair comparison, we use the results either from original papers or run publicly available implementations under the same settings.

\noindent\textbf{Implementation Details.} 
Our model is implemented in PyTorch, with the core rasterization engine extended from the 3DGS framework to support the asymmetric 2D Skew-Normal kernel. All experiments are conducted on NVIDIA RTX A6000 GPUs. In terms of computational efficiency, SNS demonstrates a training overhead comparable to SSS; for instance, in the drjohnson scene, SNS completes 30,000 iterations in 38m 4s (29.89 dB PSNR), while SSS requires 38m 7s (29.66 dB PSNR) on the same test set.

\begin{figure*}[tb]
    \centering

    \resizebox{\textwidth}{!}{%
        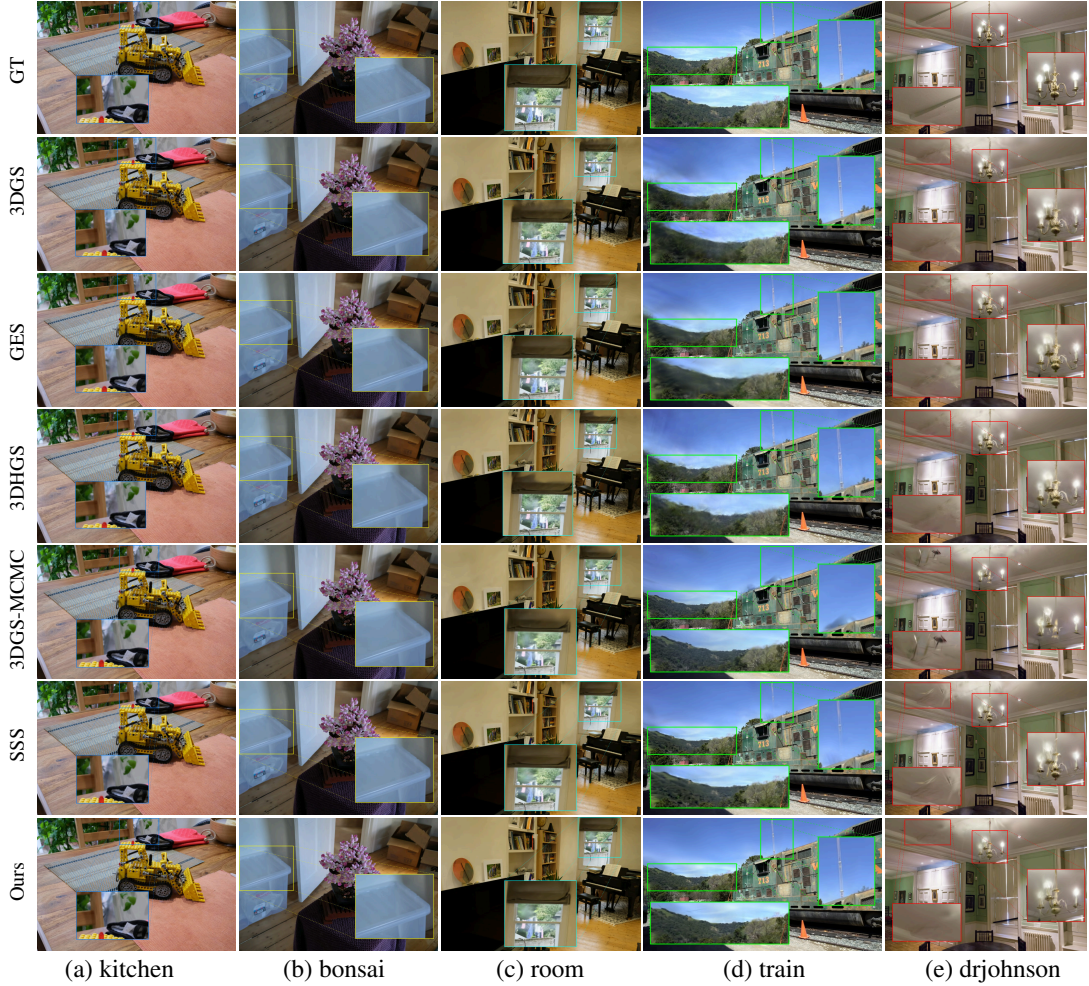
    }

    \vspace{0.1in}
    \caption{\textbf{Reconstruction quality.} We visualize five representative scenes, highlighting and zooming in on regions exhibiting the most prominent differences for comparison. \label{fig:para_effi_visual} }
    \vspace{-0.1in}   
\end{figure*}
\subsection{Results} 

\textbf{Performance Comparison.}

We report the quantitative comparison of our method with state-of-the-art methods in Tab.~\ref{tab:main}. Overall, our method achieves the best or highly competitive results across all datasets and metrics, validating the effectiveness and robustness of the proposed approach.
Our method showed the most consistent and strongest performance on PSNR, achieving 30.17 on Mip-NeRF360, 25.08 on Tanks\&Temples, and 30.30 on Deep Blending. 
The improvement is most notable on Tanks\&Temples. Compared with SSS, the strongest baseline, our method achieves higher PSNR (24.87$\rightarrow$25.08) and SSIM (0.873$\rightarrow$0.877), and lower LPIPS (0.138$\rightarrow$0.132) on Tanks\&Temples. SNS remains highly competitive on SSIM for Mip-NeRF360 and LPIPS for Deep Blending, achieving results nearly on par with the best-performing methods.

\textbf{Qualitative Results.}
Our method demonstrates better object completeness across multiple scenes in Fig. ~\ref{fig:para_effi_visual}. For locally complex objects, such as the curtain above the window in the \textit{room} scene and the tower-shaped structure in the \textit{train} scene, our method maintains the complete structure and clear boundaries. In contrast, SSS and 3DGS-MCMC exhibit noticeable missing regions in the \textit{train} scene, while GS and 3DHGS produce holes and fuzzy boundaries in the \textit{room} scene. 

In terms of texture clarity, our method more effectively preserves high-frequency details. For example, the wavy stripes on top of the transparent plastic box in the \textit{bonsai} scene, the texture and shadows on the cloth behind the chair in the \textit{kitchen} scene, and the distant trees in the \textit{train} scene are all reconstructed and rendered with clearer details. 
By contrast, the original GS produces over-smoothed reconstructions in texture-rich regions, leading to noticeable blur. Besides, in the \textit{drjohnson} scene, SNS exhibits the mildest artifacts among all compared methods. These results show that our method better handles complex regions in novel view synthesis.

\begin{table*}[t]
\caption{
    \textbf{Per-scene PSNR comparison of various reconstruction kernels.} We provide a detailed PSNR evaluation of the methods using different base kernels across the Mip-NeRF 360, Deep Blending, and Tanks\&Temples datasets. The best and second-best results are highlighted in dark red and light red, respectively.
}
\label{tab:detailed_psnr_comparison_with_different_kernels}
\centering
\setlength{\fboxsep}{1pt}
\renewcommand{\arraystretch}{1.2} 
\resizebox{\linewidth}{!}{
\begin{tabular}{l | ccccccc | cc | cc}
\toprule
\multirow{2}{*}{\textbf{Method}} & \multicolumn{7}{c|}{\textbf{Mip-NeRF 360}} & \multicolumn{2}{c|}{\textbf{Tanks\&Temples}} & \multicolumn{2}{c}{\textbf{Deep Blending}} \\
& Bicycle & Bonsai & Counter & Garden & Kitchen & Room & Stump & Train & Truck & Playroom & DrJohnson  \\
\midrule
3DGS & 25.25 & 31.98 & 28.70 & 27.41 & 30.32 & 30.63 & 26.55 & 21.09 & 25.18 & 30.04 & 28.77  \\
GES  & 24.87 & 31.97 & 28.75 & 27.07 & 31.07 & 31.17 & 26.17 & 21.73 & 24.94 & 30.29 & 29.35  \\
3D-HGS& 25.39 & 33.30 & 29.62 & 27.68 & 32.17 & 32.12 & 26.64 & 22.95 & 26.04 & 30.20 & 29.32 \\
SSS  & \second{25.68} & \second{33.50} & \second{29.87} & \second{28.09} & \second{32.43} & \second{32.57} & \second{27.17} & \second{23.32} & \second{26.41} & \second{30.47} & \second{29.66} \\
SNS (Ours) & \best{25.85} & \best{33.70} & \best{30.27} & \best{28.22} & \best{32.81} & \best{33.02} & \best{27.31} & \best{23.61} & \best{26.56} & \best{30.71} & \best{29.89} \\
\bottomrule
\end{tabular}
}
\vspace{-0.1in}
\end{table*}
\textbf{Different kernels comparison. }Table \ref{tab:detailed_psnr_comparison_with_different_kernels} provides the detailed PSNR scores evaluated with GS, GES, HGS, SSS, and our SNS. In this comparison, we treat standard 3DGS as the baseline. Across all scenes, SNS outperforms all other models by at least 0.1 dB. For example, our method outperforms SSS by 0.45 dB on the Room scene, 0.29 dB on the Train scene, and 0.24 dB on the Playroom scene. Overall, our SN formulation consistently delivers the best performance across all tested 3D scenes.

\subsection{Ablations}
Table ~\ref{tab:ablation_average} details the ablation study on individual module contributions. Using the vanilla Skew-Normal distribution as our baseline, we compare three configurations:
\textbf{(1) $k$-decomposition}: Decomposes the skew vector $k$ into magnitude and direction, without alternating optimization.
\textbf{(2) Alternating optimization}: Alternately optimizes $(R, S)$ and $k$, without $k$-decomposition.
\textbf{(3) Full model}: Incorporates both $k$-decomposition and alternating optimization.
Full experimental results are available in Appendix \ref{sec:detailed_results}.

Both modules contribute to the final performance through distinct mechanisms. Alternating optimization yields consistent improvements in pixel-level reconstruction, increasing mean PSNR from $30.06$ to $30.14$ on Mip-NeRF 360 and from $30.16$ to $30.30$ on Deep Blending. This indicates that alternating updates for $(R, S)$ and $k$ effectively decouple geometric parameters from skew parameters, stabilizing the optimization process. Meanwhile, $k$-decomposition consistently enhances SSIM and LPIPS across all datasets, suggesting that decomposing the skew vector into magnitude and direction provides a superior parameterization that improves optimization dynamics. The full model delivers the most significant gains in PSNR, while maintaining highly competitive LPIPS and SSIM performance across scenes.
Thus, these results suggest that the two optimization components provide complementary but moderate gains. The full model achieves the best average PSNR across all datasets, while individual variants may slightly outperform it on perceptual metrics in some cases, indicating that the modules mainly improve reconstruction fidelity and stability rather than uniformly optimizing every metric.

\begin{table*}[t]
\caption{
    \textbf{Summary ablation across datasets.} We present the average performance of our Full SNS compared with its ablated versions across three main datasets. The best and second-best results are highlighted in dark red and light red, respectively.
    \label{tab:ablation_average}
}
\setlength{\fboxsep}{1pt} 
\renewcommand{\arraystretch}{1.2}
\resizebox{\linewidth}{!}{
\begin{tabular}{l|ccc|ccc|ccc}
\toprule
    Dataset & \multicolumn{3}{c|}{Mip-NeRF360}  & \multicolumn{3}{c|}{Tanks\&Temples} & \multicolumn{3}{c}{Deep Blending}\\
    Model Variants & PSNR $\uparrow$   & SSIM$\uparrow$   & LPIPS$\downarrow$ & PSNR $\uparrow$   & SSIM$\uparrow$   & LPIPS$\downarrow$ & PSNR $\uparrow$   & SSIM$\uparrow$   & LPIPS$\downarrow$ \\
    \midrule 
    Vanilla SNS & 30.0643 & 0.8942 & \second{0.1443} & \second{25.0388} & 0.8765 & 0.1324 & 30.1581 & 0.9097 & 0.2448 \\
    $k$-decomposition & 30.1204 & \second{0.8950} & \best{0.1442} & 25.0090 & 0.8768 & \best{0.1319} & 30.1185 & 0.9118 & \second{0.2441} \\
    Alternating Optimization & \second{30.1390} & 0.8949 & 0.1445 & 25.0328 & \best{0.8776} & \second{0.1322} & \second{30.2969} & \best{0.9122} & \best{0.2436} \\
    \midrule 
    Full SNS& \best{30.1700} & \best{0.8954} & 0.1445 & \best{25.0839} & \second{0.8775} & 0.1325 & \best{30.3034} & \second{0.9119} & 0.2444 \\
    \bottomrule
\end{tabular}
}
\vspace{-0.1in}
\end{table*}
\section{Conclusion}
We propose Skew-Normal Splatting, which introduces controllable skewness into Gaussian Splatting to better represent asymmetric geometries, sharp boundaries, and one-sided structures. SNS preserves analytical projection for efficient splatting and further improves optimization stability through decoupled parameterization and block-wise updates. Experiments show consistent improvements over existing Gaussian and non-Gaussian kernels.
\newline

{
\bibliographystyle{unsrt}
\bibliography{main}
}

\clearpage
\appendix
\begin{center}
    {\Large \bfseries Appendix}
\end{center}

\section{Mathematical Proof}
In the implementation of SNS, the kernel is defined using the unnormalized probability density function (PDF) of the Skew-Normal distribution. Although this differs from the standard Skew-Normal (SN) distribution by a normalization coefficient, they share identical properties regarding affine transformations and geometric characteristics. For clarity and convenience, the following derivations are conducted within the framework of the standard SN distribution.
\subsection{Closure under Affine Transformations}
\label{sec:closure}
Let $X \sim SN_d(\mu, \Omega, \alpha)$ be a $d$-dimensional Skew-Normal random variable. Its stochastic representation~\cite{Azzalini_1999} is given by:
\begin{equation}
\label{eq:stochastic representation}
    X = \mu + \omega Z \mid U > 0, \quad
    \begin{pmatrix} U \\ Z \end{pmatrix} \sim
    N_{1+d}\!\left(
    \mathbf{0},
    \begin{pmatrix}
    1 & \delta^\top \\
    \delta & \bar{\Omega}
    \end{pmatrix}
    \right),
\end{equation}
where $\omega = \left( \Omega \odot \mathbf{I}_d \right)^{1/2}$, $\bar{\Omega} = \omega^{-1}\Omega\omega^{-1}$, and $\delta = \frac{\bar{\Omega}\alpha}{\sqrt{1+\alpha^\top\bar{\Omega}\alpha}}$.

Consider an affine transformation $Y = TX + b$, where $T \in \mathbb{R}^{h \times d}$, $\text{rank}(T) = h \leq d$ and $b \in \mathbb{R}^h$. Substituting the representation of $X$ yields: $Y = (T\mu + b) + T\omega Z \mid U > 0$.

To show that $Y$ is also Skew-Normal, we express it in the standard stochastic representation. Let $\mu_Y = T\mu + b$ and $W = T\omega Z$. Since $(U, Z^\top)^\top$ is multivariate normal, the joint distribution of $(U, W^\top)^\top$ remains normal with zero mean. The variance of $W$ and the covariance between $U$ and $W$ are given by:
\begin{equation}
    \begin{aligned}
        &\text{Var}(W) = \Omega_Y = T\omega \bar{\Omega} \omega T^\top = T\Omega T^\top,\\
        &\text{Cov}(U, W) = \text{Cov}(U, Z)\omega T^\top = \delta^\top \omega T^\top.
    \end{aligned}
\end{equation}

Defining $\omega_Y = \left( \Omega_Y \odot \mathbf{I}_h \right)^{1/2}$ and $Z_Y = \omega_Y^{-1} W$, we can rewrite $Y = \mu_Y + \omega_Y Z_Y \mid U > 0$, while $\text{Var}(Z_Y) = \omega_Y^{-1} \Omega_Y \omega_Y^{-1} = \bar{\Omega}_Y$, and $\text{Cov}(U, Z_Y) = \text{Cov}(U, W)\omega_Y^{-1} = \delta^\top \omega T^\top \omega_Y^{-1}$. Letting $\delta_Y = \omega_Y^{-1} T \omega \delta$, we finally obtain the required block-covariance structure:
\begin{equation}
    Y = \mu_Y + \omega_Y Z_Y \mid U > 0, \quad
    \begin{pmatrix} U \\ Z_Y \end{pmatrix} \sim
    N_{1+h}\!\left(
    \mathbf{0},
    \begin{pmatrix}
    1 & \delta_Y^\top \\
    \delta_Y & \bar{\Omega}_Y
    \end{pmatrix}
    \right).
\end{equation}

Since the conditioning event $U > 0$ is unchanged and $(U, Z_Y^\top)^\top$ satisfies the standard joint normal structure, we conclude that $Y$ follows an $h$-dimensional SND.
    
\subsection{2D Kernel Parameters}

Let $X \sim \text{SN}_3(\mu, \Omega, \alpha)$ represent the 3D Skew-Normal kernel, and consider a linear projection: $Y = A X$, where $A \in \mathbb{R}^{2 \times 3}$. Building upon the property derived in \ref{sec:closure}, we can simply write  $Y \sim SN_2(\mu_{Y}, \Omega_{Y}, \alpha_{Y})$, with $\mu_{Y} = A \mu$, $\Omega_{Y} = A \Omega A^\top$, and $\delta_Y = \omega_Y^{-1} A \omega \delta$.

Recall that $\delta_Y = \frac{\bar{\Omega}_Y \alpha_Y}{\sqrt{1 + \alpha_Y^\top \bar{\Omega}_Y\alpha_Y}}$ and $1-\delta_Y ^\top \bar\Omega_Y^{-1} \delta_Y = \frac{1}{1 + \alpha_Y^\top \bar{\Omega}_Y\alpha_Y}$. We can write:
\begin{equation}
    \label{eq:alpha_Y_1}
    \begin{aligned}
    \alpha_Y = \sqrt{1 + \alpha_Y^\top \bar{\Omega}_Y\alpha_Y} \cdot \bar\Omega_Y^{-1} \delta_Y = \frac{\bar\Omega_Y^{-1} \delta_Y}{\sqrt{1-\delta_Y ^\top \bar\Omega_Y^{-1} \delta_Y}} = \frac{\omega_{Y} \Omega_{Y}^{-1} A \omega \delta}{\sqrt{1 - (A \omega \delta)^\top\Omega_{Y}^{-1} A \omega \delta}},
\end{aligned}
\end{equation}

where $\omega_Y = (\Omega_Y \odot \mathbf{I}_2)^{1/2}$, $\omega = (\Omega \odot \mathbf{I}_3)^{1/2}$, and $\delta = \frac{\bar{\Omega}\alpha}{\sqrt{1+\alpha^\top\bar{\Omega}\alpha}}$.

In our formulation, $\alpha$ is explicitly parameterized as $\alpha = \omega R S^{-1} k$, where $\Omega = RS S^\top R^\top$. Substituting this structural definition into the equations above simplifies the expressions:
\begin{equation}\label{eq:omega-delta}
    \begin{aligned}
        \omega\delta &= \frac{\omega\bar{\Omega}\omega R S^{-1} k}{\sqrt{1+k^\top S^{-1} R^\top \omega\bar{\Omega}\omega R S^{-1} k}} = \frac{\Omega R S^{-1} k}{\sqrt{1+k^\top S^{-1} R^\top \Omega R S^{-1} k}} \\
        & = \frac{RS S^\top R^\top R S^{-1} k}{\sqrt{1+k^\top S^{-1} R^\top RS S^\top R^\top R S^{-1} k}} = \frac{RSk}{\sqrt{1+k^\top k}}.
    \end{aligned}
\end{equation}

Applying the transformation Eq.(\ref{eq:omega-delta}) to Eq.(\ref{eq:alpha_Y_1}) gives:
\begin{equation}
    \alpha_Y = \frac{\omega_{Y}\Omega_Y^{-1} A R S k}{\sqrt{1 + k^\top k - (A R S k)^\top (\Omega_Y)^{-1} A R S k}}.
\end{equation}

\subsection{The Mean of the Skew-Normal Distribution}
Unlike the multivariate normal distribution, the location parameter $\mu$ of $X \sim SN_d(\mu, \Omega, \alpha)$ does not equal its exact mean. The true mean, $\mathbb{E}[X]$, can be analytically derived using the stochastic representation introduced in Eq.(\ref{eq:stochastic representation}). Taking the expectation of $X = \mu + \omega Z \mid U > 0$, we have:
\begin{equation}
\begin{aligned}
    \mathbb{E}[X] &= \mu + \omega \mathbb{E}[Z \mid U > 0] \\
    &= \mu + \omega \mathbb{E}[ \mathbb{E}[ Z \mid U] \mid U > 0] \\
    &= \mu + \omega \left([\mathbb{E}[Z-U\delta] + \mathbb{E}[ U\delta \mid U > 0]\right) \\
    &= \mu + \mathbb{E}[ U \mid U > 0] \cdot \omega\delta  \\
    &= \mu + \sqrt{\frac{2}{\pi}} \cdot \omega \delta.
\end{aligned}
\end{equation}

If $\alpha = \omega R S^{-1} k$, associated with Eq.(\ref{eq:omega-delta}), we have:
\begin{equation}
    \mathbb{E}[X] = \mu + \sqrt{\frac{2}{\pi}} \cdot \frac{RSk}{\sqrt{1+k^\top k}}
\end{equation}

\section{Backward Pass}
\label{sec:backward_pass}
The learnable parameters for each component in SNS include the location $\mu$, scale matrix $\Omega$($i.e.$ $S$ and $R$), color $c$, opacity $o$, skew $k$, and the key gradients are $\frac{\partial L}{\partial \mu}$, $\frac{\partial L}{\partial S}$, $\frac{\partial L}{\partial R}$, $\frac{\partial L}{\partial c}$, $\frac{\partial L}{\partial o}$, $\frac{\partial L}{\partial k}$, where $L$ is the loss. 
$L$ is calculated between the ground-truth pixel colors and rendered colors. The rendered pixel colors are:
\begin{equation}
    C(u) = \sum_{i=1}^{N} c_i a_i(u) \prod_{j=1}^{i-1} \left(1 - a_j(u)\right),
    \label{eq:rendering1}
\end{equation}
where $a(u) = o \cdot SN^{2D}(u)$ and u denotes pixel coordinate in 2D images.

Previous methods, such as 3DGS and SSS, compute the gradients with respect to the shape parameters ($R$, $S$) by first calculating $\frac{\partial L}{\partial (\Omega^{-1})}$ and subsequently backpropagating to obtain $\frac{\partial L}{\partial R}$ and $\frac{\partial L}{\partial S}$. However, recalling the definitions in Eq.~\eqref{eq:2D_krenl} and Eq.~\eqref{eq:affine_transform_params}, computing $\frac{\partial L}{\partial \Omega}$ directly proves cumbersome. The parameters ($R$, $S$) couple intricately within both $ARSk$ and $(\Omega^{2D})^{-1}$, leading to high computational complexity. To mitigate this, we propose an alternative differentiation scheme. By introducing an intermediate variable $Q = ARS$, we reformulate Eq.~\eqref{eq:2D_krenl} and Eq.~\eqref{eq:affine_transform_params} as follows:

\begin{equation}
\label{eq:2D_krenl_rewritting}
    SN^{2D}(u) = G_2(u - \mu^{2D}; QQ^\top) \, \Phi \left( (m^{2D})^\top (u - \mu^{2D}) \right),
\end{equation}
where 
\begin{equation*}
    m^{2D} = \frac{(QQ^\top)^{-1}Qk}{\sqrt{1 + k^\top k - k^\top Q^\top (QQ^\top)^{-1}Qk}}.
\end{equation*}

Applying the chain rule, the gradients are given by:
\begin{equation}
\label{eq:grad}
\begin{aligned}
    \frac{\partial L}{\partial c} &= \frac{\partial L}{\partial C} \cdot \frac{\partial C}{\partial c}, \\
    \frac{\partial L}{\partial o} &= \frac{\partial L}{\partial C} \cdot \frac{\partial C}{\partial a} \cdot \frac{\partial a}{\partial o}, \\
    \frac{\partial L}{\partial \mu} &= \frac{\partial L}{\partial C} \cdot \frac{\partial C}{\partial a} \cdot \frac{\partial a}{\partial SN^{2D}} \cdot \frac{\partial SN^{2D}}{\partial \mu^{2D}} \cdot \frac{\partial\mu^{2D}}{\partial \mu}, \\
    \frac{\partial L}{\partial k} &= \frac{\partial L}{\partial C} \cdot \frac{\partial C}{\partial a} \cdot \frac{\partial a}{\partial SN^{2D}} \cdot \frac{\partial SN^{2D}}{\partial k}, \\
    \frac{\partial L}{\partial R} &= \frac{\partial L}{\partial C} \cdot \frac{\partial C}{\partial a} \cdot \frac{\partial a}{\partial SN^{2D}} \cdot \frac{\partial SN^{2D}}{\partial Q} \cdot \frac{\partial Q}{\partial R}, \\
    \frac{\partial L}{\partial S} &= \frac{\partial L}{\partial C} \cdot \frac{\partial C}{\partial a} \cdot \frac{\partial a}{\partial SN^{2D}} \cdot \frac{\partial SN^{2D}}{\partial Q} \cdot \frac{\partial Q}{\partial S},
\end{aligned}
\end{equation}
where $C$ and $a$ are defined in Eq.~\eqref{eq:rendering1}. 
Compared to the standard gradient computation in 3DGS, our formulation solely requires deriving the partial derivatives of $SN^{2D}$ with respect to $\mu^{2D}$, $k$, and $Q$; the gradients for all other standard components remain unchanged and are therefore omitted. The final analytical expressions for these specific derivatives are presented in Eq.~\eqref{eq:SN-mu-final}, Eq.~\eqref{eq:SN-k-final}, and Eq.~\eqref{eq:SN-Q-final}, respectively. For notational simplicity, we denote $m = m^{2D}$, $\Delta = u - \mu^{2D}$, and $D = \sqrt{1 + k^\top k - k^\top Q^\top (\Omega^{2D})^{-1} Q k}$. Also, we define the following auxiliary variables:

\begin{equation}
\label{eq:lambda-Delta}
    \begin{aligned}
        \boldsymbol{c}_{\Delta} &= \Phi \cdot G_2 \cdot  (\Omega^{2D})^{-1}\Delta, \\
        \boldsymbol{M}_{\Delta} &= \Phi \cdot G_2 \cdot \left( (\Omega^{2D})^{-1} \Delta\right)\left( (\Omega^{2D})^{-1} \Delta\right)^\top, \\
        s_{\Delta} &= \frac{G_2}{\sqrt{2\pi}} \exp\left(-\frac{1}{2}(m^\top \Delta)^2\right), \\
        \boldsymbol{p}_{\Delta} &= \frac{G_2}{\sqrt{2\pi}} \exp\left(-\frac{1}{2}(m^\top\Delta)^2\right) \cdot (\Omega^{2D})^{-1}\Delta,
    \end{aligned}
\end{equation}
where $s_{\Delta}\in \mathbb{R}$, $\,\boldsymbol{c}_{\Delta}, \boldsymbol{p}_{\Delta} \in \mathbb{R}^2$ and $\boldsymbol{M}_{\Delta} \in \mathbb{R}^{2 \times 2}$.

\noindent\textbf{Gradient with respect to $\mu^{2D}$.} 

First, we derive the gradient with respect to the mean vector $\mu^{2D}$. Applying the product rule, we obtain:
\begin{equation}
    \frac{\partial SN^{2D}}{\partial \mu^{2D}} = \Phi \cdot \frac{\partial G_2}{\partial \mu^{2D}} + G_2 \cdot \frac{\partial \Phi}{\partial \mu^{2D}}.
\end{equation}

Given $\Delta = u - \mu^{2D}$, the inner derivative is $\frac{\partial \Delta}{\partial \mu^{2D}} = -I$. So the derivative of the Gaussian component $G_2$ is:
\begin{equation}
    \frac{\partial G_2}{\partial \mu^{2D}} = G_2 (\Omega^{2D})^{-1} (u - \mu^{2D}) = G_2 (\Omega^{2D})^{-1} \Delta.
\end{equation}

For the cumulative distribution function (CDF) component $\Phi(m^\top \Delta)$, the chain rule yields:
\begin{equation}
    \frac{\partial \Phi(m^\top \Delta)}{\partial \mu^{2D}} = -\frac{1}{\sqrt{2\pi}} \exp\left(-\frac{1}{2}(m^\top \Delta)^2\right) m.
\end{equation}

Substituting these partial derivatives back into the product rule and utilizing the definitions in Eq.~\eqref{eq:lambda-Delta}, we arrive at the intermediate expression:
\begin{equation}
\label{eq:SN-mu-final}
    \begin{aligned}
        \frac{\partial SN^{2D}}{\partial \mu^{2D}} &= \Phi \cdot G_2 (\Omega^{2D})^{-1} \Delta - \frac{G_2}{\sqrt{2\pi}} \exp\left(-\frac{1}{2}(m^\top \Delta)^2\right) m \\
        &= \boldsymbol{c}_{\Delta} - s_{\Delta} \cdot m.
    \end{aligned}
\end{equation}

\noindent\textbf{Gradient with respect to $k$.} 

Next, we derive the gradient with respect to $k$:
\begin{equation}
\label{eq:SN-k}
    \begin{aligned}
        \frac{\partial SN^{2D}}{\partial k} &= G_2 \cdot \frac{\partial \Phi}{\partial (m^\top \Delta)} \cdot \frac{\partial (m^\top \Delta)}{\partial k} \\
        &= \frac{G_2}{\sqrt{2\pi}} \exp\left(-\frac{1}{2}(m^\top\Delta)^2\right) \cdot \frac{\partial (m^\top \Delta)}{\partial k},
    \end{aligned}
\end{equation}
where
\begin{equation}
\label{eq:m_Delta-k}
    \begin{aligned}
        \frac{\partial(m^\top \Delta)}{\partial k} =& \frac{1}{D} \cdot \frac{\partial \left(\Delta^\top (\Omega^{2D})^{-1}Qk\right)}{\partial k} \\
        &- \frac{\Delta^\top (\Omega^{2D})^{-1}Qk}{2D^3} \cdot \frac{\partial \left(k^\top k - k^\top Q^\top (\Omega^{2D})^{-1} Q k\right)}{\partial k} \\
        =& \frac{Q^\top (\Omega^{2D})^{-1} \Delta}{D} - \frac{\Delta^\top(\Omega^{2D})^{-1}Qk}{D^3} \cdot \left(k - Q^\top (\Omega^{2D})^{-1} Q k\right) \\
        =& \frac{Q^\top (\Omega^{2D})^{-1} \Delta}{D} + \frac{(\Delta^\top m) \cdot Q^\top m }{D} - \frac{(\Delta^\top m) \cdot k}{D^2} \\
        =& \frac{1}{D} Q^\top \left( (\Omega^{2D})^{-1} \Delta + (\Delta^\top m) \cdot m \right) - \frac{1}{D^2} \cdot (\Delta^\top m) \cdot k.
    \end{aligned}
\end{equation}

Combining Eq.~\eqref{eq:lambda-Delta}, Eq.~\eqref{eq:SN-k}, and Eq.~\eqref{eq:m_Delta-k}, we obtain:
\begin{equation}
\label{eq:SN-k-final}
    \begin{aligned}
        \frac{\partial SN^{2D}}{\partial k} &= \frac{1}{D} Q^\top \left( \boldsymbol{p}_{\Delta} + (\boldsymbol{p}_{\Delta}^\top \Omega^{2D} m) \cdot m \right) - \frac{1}{D^2} \cdot (\boldsymbol{p}_{\Delta}^\top \Omega^{2D} m) \cdot k.
    \end{aligned}
\end{equation}

\noindent\textbf{Gradient with respect to $Q$.} 

To compute $\frac{\partial SN^{2D}}{\partial Q}$, we apply the product rule to the respective terms:
\begin{equation}
    \begin{aligned}
        \frac{\partial SN^{2D}}{\partial Q} = G_2 \cdot \frac{\partial \Phi}{\partial (m^\top \Delta)} \cdot \frac{\partial (m^\top \Delta)}{\partial Q} + \Phi \cdot \frac{\partial G_2}{\partial Q}.
    \end{aligned}
\end{equation}

First, we calculate the gradient of the 2D Gaussian $G_2$ with respect to $Q$. The differential of $G_2$ can be expressed using the trace operator $\text{Tr}(\cdot)$:
\begin{equation}
\label{eq:G2-Q-deriv}
    \begin{aligned}
        dG_2 &= G_2 \cdot \left(- \frac{1}{2} \Delta^\top d(\Omega^{2D})^{-1} \Delta\right) \\
        &= \frac{1}{2} G_2 \text{Tr}\left( \Delta^\top(\Omega^{2D})^{-1}(Q dQ^\top + dQ Q^\top)(\Omega^{2D})^{-1}\Delta \right) \\
        &= \text{Tr}\left( G_2 \cdot Q^\top (\Omega^{2D})^{-1} \Delta\Delta^\top (\Omega^{2D})^{-1}dQ \right).
    \end{aligned}
\end{equation}
By the matrix calculus identity where the differential $df = \text{Tr}(P^\top dQ)$ implies the gradient $\frac{\partial f}{\partial Q} = P$, we extract the derivative with respect to $Q$:
\begin{equation}
\label{eq:G2-Q-final}
    \frac{\partial G_2}{\partial Q} = G_2 \left( (\Omega^{2D})^{-1} \Delta \Delta^\top (\Omega^{2D})^{-1} \right) Q.
\end{equation}

Next, we compute $\frac{\partial (m^\top \Delta)}{\partial Q}$. Let $N = \Delta^\top (\Omega^{2D})^{-1} Q k$, which implies $m^\top \Delta = \frac{N}{D}$. By the quotient rule, its differential is:
\begin{equation}
\label{eq:diff-m_Delta}
    d(m^\top \Delta) = \frac{1}{D} dN - \frac{N}{D^2} dD = \frac{1}{D} dN - \frac{m^\top \Delta}{D} dD.
\end{equation}

For $dN$, evaluating the differential and substituting $(\Omega^{2D})^{-1} Q k = D m$, we have:
\begin{equation}
    \begin{aligned}
        dN &= \Delta^\top d\left((\Omega^{2D})^{-1}\right) Q k + \Delta^\top (\Omega^{2D})^{-1} dQ \, k \\
        &= -\Delta^\top (\Omega^{2D})^{-1} (dQ \, Q^\top + Q \, dQ^\top) (\Omega^{2D})^{-1} Q k + \Delta^\top (\Omega^{2D})^{-1} dQ \, k \\
        &= \text{Tr}\left( k \Delta^\top (\Omega^{2D})^{-1} dQ - Q^\top D m \Delta^\top (\Omega^{2D})^{-1} dQ - D m \Delta^\top (\Omega^{2D})^{-1} Q dQ^\top \right) \\
        &= \text{Tr}\left( \left[ (\Omega^{2D})^{-1} \Delta k^\top - D (\Omega^{2D})^{-1} \Delta m^\top Q - D m \Delta^\top (\Omega^{2D})^{-1} Q \right]^\top dQ \right).
    \end{aligned}
\end{equation}
This establishes the gradient of $N$:
\begin{equation}
\label{eq:N-Q}
    \frac{\partial N}{\partial Q} = (\Omega^{2D})^{-1} \Delta k^\top - D \left( (\Omega^{2D})^{-1} \Delta m^\top + m \Delta^\top (\Omega^{2D})^{-1} \right) Q.
\end{equation}

Similarly, we derive variable $D$'s differential:
\begin{equation}
    \begin{aligned}
        dD &= - \frac{1}{2D} \cdot d\left(k^\top Q^\top (\Omega^{2D})^{-1} Qk\right) \\
        &= - \frac{1}{D} \cdot \left( k^\top Q^\top (\Omega^{2D})^{-1} dQ \, k - k^\top Q^\top (\Omega^{2D})^{-1} dQ \, Q^\top (\Omega^{2D})^{-1} Q k \right) \\
        &= - \frac{1}{D} \cdot \left( D m^\top dQ \, k - D^2 m^\top dQ \, Q^\top m \right) \\
        &= \text{Tr}\left( \left(-mk^\top + D \cdot mm^\top Q\right)^\top dQ \right),
    \end{aligned}
\end{equation}
which yields: 
\begin{equation}
\label{eq:D-Q}
    \frac{\partial D}{\partial Q } = -mk^\top + D \cdot mm^\top Q.
\end{equation}

Combining Eq.~\eqref{eq:diff-m_Delta}, Eq.~\eqref{eq:N-Q}, and Eq.~\eqref{eq:D-Q}, we assemble $\frac{\partial (m^\top \Delta)}{\partial Q}$:
\begin{equation}
\label{eq:m_Delta-Q}
    \begin{aligned}
        \frac{\partial (m^\top \Delta)}{\partial Q} =& \frac{1}{D} \left( (\Omega^{2D})^{-1} \Delta + (m^\top \Delta) m \right) k^\top \\
        &- \left( (\Omega^{2D})^{-1} \Delta m^\top + m \Delta^\top (\Omega^{2D})^{-1} + (m^\top \Delta) m m^\top \right) Q.
    \end{aligned}
\end{equation}

Finally, recalling that $G_2 \cdot \frac{\partial \Phi}{\partial (m^\top \Delta)} = \frac{G_2}{\sqrt{2\pi}} \exp\left(-\frac{1}{2}(m^\top \Delta)^2\right)$, we multiply Eq.~\eqref{eq:m_Delta-Q} by this term. Substituting the previously defined variables $\boldsymbol{p}_{\Delta}$ and $M_\Delta$, the full analytical gradient $\frac{\partial SN^{2D}}{\partial Q}$ is derived as:
\begin{equation}
\label{eq:SN-Q-final}
    \begin{aligned}
        \frac{\partial SN^{2D}}{\partial Q} =& \left(\boldsymbol{M}_{\Delta} - \boldsymbol{p}_{\Delta} m^\top - m \boldsymbol{p}_{\Delta}^\top - (\boldsymbol{p}_{\Delta}^\top \Omega^{2D} m) \cdot m m^\top \right) Q \\
        &+ \frac{1}{D} \left( \boldsymbol{p}_{\Delta} + (\boldsymbol{p}_{\Delta}^\top \Omega^{2D} m) \cdot m \right) k^\top.
    \end{aligned}
\end{equation}
Following the standard 3DGS architecture, the backward pass consists of rendering and preprocessing stages. In standard 3DGS, the rendering stage first computes gradients with respect to 2D image-space variables (e.g., $\frac{\partial L}{\partial \mu^{2D}}$ and $\frac{\partial L}{\partial (\Omega^{2D})^{-1}}$), and the preprocessing stage uses these intermediate gradients to compute the gradients for 3D learnable parameters. 

By contrast, the SNS rendering stage directly evaluates the shared auxiliary terms ($\boldsymbol{c}_{\Delta}, \boldsymbol{M}_{\Delta}, s_{\Delta}, \boldsymbol{p}_{\Delta}$), which are intermediate algebraic quantities rather than derivatives. Next, the preprocessing stage leverages these precomputed terms to efficiently evaluate the exact gradients of the 3D parameters using our derived closed-form expressions (Eq.~\eqref{eq:SN-mu-final}, Eq.~\eqref{eq:SN-k-final}, and Eq.~\eqref{eq:SN-Q-final}). By algebraically restructuring the complex gradient derivation into clear and concise computations, this formulation maximizes variable reuse and ensures a smooth and efficient computational flow.

\section{Detailed results}
\label{sec:detailed_results}
In this section, we present the detailed results. Specifically, Table \ref{tab:final_comparison} reports the complete PSNR, SSIM, and LPIPS metrics for our full SNS model and its ablations across all evaluated scenes.
\begin{table*}[t]
\caption{
{\bf Per-scene ablation details. } This table presents the performance of our Full SNS compared with its ablated versions. The Full SNS is composed of a baseline combined with k-decomposition and Alternating Optimization. We report PSNR, SSIM, and LPIPS for each scene and the average per dataset.
}
\centering
\resizebox{\linewidth}{!}{
\setlength{\tabcolsep}{5pt}
\renewcommand{\arraystretch}{1.2}
\begin{tabular}{@{}c l cccc@{}}
\toprule
& & \textbf{Vanilla SNS} & \textbf{$k$-decomposition} & \textbf{Alternating Optimization} & \textbf{Full SNS} \\
\textbf{Dataset} & \textbf{Scene}
& {\tiny PSNR$\uparrow$ / SSIM$\uparrow$ / LPIPS$\downarrow$}
& {\tiny PSNR$\uparrow$ / SSIM$\uparrow$ / LPIPS$\downarrow$}
& {\tiny PSNR$\uparrow$ / SSIM$\uparrow$ / LPIPS$\downarrow$}
& {\tiny PSNR$\uparrow$ / SSIM$\uparrow$ / LPIPS$\downarrow$} \\
\midrule

\multirow{8}{*}{\rotatebox{90}{Mip-NeRF 360}}
& Counter
& 30.1129 / 0.9270 / 0.1534
& 30.2460 / 0.9283 / 0.1531
& 30.1116 / 0.9277 / 0.1540
& 30.2749 / 0.9282 / 0.1537 \\

& Bicycle
& 25.7461 / 0.8004 / 0.1719
& 25.8033 / 0.8008 / 0.1726
& 25.8162 / 0.8011 / 0.1724
& 25.8510 / 0.8024 / 0.1722 \\

& Stump
& 27.2857 / 0.8144 / 0.1739
& 27.3576 / 0.8151 / 0.1744
& 27.3833 / 0.8144 / 0.1756
& 27.3130 / 0.8149 / 0.1755 \\

& Garden
& 28.1096 / 0.8828 / 0.0909
& 28.1546 / 0.8832 / 0.0905
& 28.1348 / 0.8830 / 0.0910
& 28.2243 / 0.8837 / 0.0908 \\

& Bonsai
& 33.6172 / 0.9551 / 0.1514
& 33.7117 / 0.9563 / 0.1505
& 33.7363 / 0.9562 / 0.1509
& 33.7026 / 0.9568 / 0.1514 \\

& Room
& 33.0042 / 0.9392 / 0.1662
& 32.9194 / 0.9397 / 0.1660
& 33.0393 / 0.9401 / 0.1658
& 33.0163 / 0.9400 / 0.1665 \\

& Kitchen
& 32.5742 / 0.9402 / 0.1022
& 32.6499 / 0.9415 / 0.1020
& 32.7514 / 0.9419 / 0.1017
& 32.8082 / 0.9420 / 0.1016 \\

\cmidrule{2-6}
& Average
& 30.0643 / 0.8942 / 0.1443
& 30.1204 / 0.8950 / 0.1442
& 30.1390 / 0.8949 / 0.1445
& 30.1700 / 0.8954 / 0.1445 \\

\midrule

\multirow{3}{*}{\rotatebox[origin=c]{90}{
    \begin{tabular}{@{}c@{}}
        Deep \\ 
        Blending 
    \end{tabular}
}}
& Playroom
& 30.4695 / 0.9099 / 0.2446
& 30.3995 / 0.9144 / 0.2425
& 30.6745 / 0.9153 / 0.2410
& 30.7130 / 0.9148 / 0.2425 \\

& DrJohnson
& 29.8467 / 0.9095 / 0.2450
& 29.8375 / 0.9091 / 0.2456
& 29.9192 / 0.9091 / 0.2461
& 29.8938 / 0.9089 / 0.2462 \\

\cmidrule{2-6}
& Average
& 30.1581 / 0.9097 / 0.2448
& 30.1185 / 0.9118 / 0.2441
& 30.2969 / 0.9122 / 0.2436
& 30.3034 / 0.9119 / 0.2444 \\

\midrule
\multirow{3}{*}{\rotatebox[origin=c]{90}{
    \begin{tabular}{@{}c@{}}
        Tanks \& \\ 
        Temples 
    \end{tabular}
}}
& Train
& 23.5475 / 0.8548 / 0.1604
& 23.5138 / 0.8548 / 0.1614
& 23.5132 / 0.8564 / 0.1606
& 23.6065 / 0.8561 / 0.1615 \\

& Truck
& 26.5301 / 0.8981 / 0.1044
& 26.5041 / 0.8987 / 0.1024
& 26.5523 / 0.8987 / 0.1037
& 26.5613 / 0.8989 / 0.1035 \\

\cmidrule{2-6}
& Average
& 25.0388 / 0.8765 / 0.1324
& 25.0090 / 0.8768 / 0.1319
& 25.0328 / 0.8776 / 0.1322
& 25.0839 / 0.8775 / 0.1325 \\

\bottomrule
\end{tabular}
}
\label{tab:final_comparison}
\end{table*}

\section{Alternating Optimization Strategy}
\begin{algorithm}[t]
\caption{Alternating Optimization for Primitive Shape}
\label{alg:bcd}
\begin{algorithmic}[1]
\Require Current iteration step $t$, warm-up threshold $T_{start}$, cycle length $C$, base phase duration $C_{base}$.
\Require Base shape parameters $\Theta_{base}$, skewness parameters $\Theta_{skew}$.

\State Compute loss $\mathcal{L}$ and initial gradients for all parameters
\If{$t > T_{start}$} \Comment{Begin Alternating Optimization}
    \State $step_{cycle} \gets t \pmod C$
    \If{$step_{cycle} < C_{base}$}
        \State Freeze $\Theta_{skew}$ \Comment{Zero gradients \& LR for skewness}
        \State Unfreeze $\Theta_{base}$ 
    \Else
        \State Freeze $\Theta_{base}$ \Comment{Zero gradients \& LR for base shape}
        \State Unfreeze $\Theta_{skew}$
    \EndIf
\Else
    \State Unfreeze all $\Theta_{base}$ and $\Theta_{skew}$ \Comment{Joint optimization}
\EndIf
\State Apply optimizer step (e.g., Adam/SGHMC) to update unfrozen parameters
\end{algorithmic}
\end{algorithm}
Jointly optimizing the base geometric attributes (rotation and scaling) with the skewness parameters can lead to optimization instability due to their highly coupled influence on the primitive's spatial footprint. To mitigate this, we introduce an Alternating Optimization strategy inspired by Block Coordinate Descent (BCD). 

As outlined in Algorithm \ref{alg:bcd}, we partition the shape-related parameters into two disjoint groups: the base shape parameters $\Theta_{base} = \{R, S\}$ and the skewness parameters $\Theta_{skew} = \{d_{k}, m_{k}\}$. During the initial warm-up phase (before iteration $T_{start}$), all parameters are optimized jointly to establish a reasonable geometry. Once the threshold $T_{start}$ is reached, we interleave the optimization of $\Theta_{base}$ and $\Theta_{skew}$ within a fixed cycle length $C$. In each cycle, we freeze the skewness to refine the base shape for $C_{base}$ steps, and subsequently freeze the base shape to optimize the skewness for the remaining steps. This decoupled optimization stabilizes the gradient descent and prevents degenerate primitive shapes.

\section{Limitations and future work.}
While SNS improves the reconstruction of asymmetric geometries and sharp boundaries, it introduces certain limitations. First, evaluating the Skew-Normal distribution requires computing the Cumulative Distribution Function (CDF).  In our current implementation, we directly employ the highly accurate numerical solution (the \textbf{erf} function) to ensure mathematical precision. While faster analytic approximations exist, relying on the precise \textbf{erf} incurs moderate penalties on rendering speed. Second, as SNS primarily enhances geometric representation, it still shares the limitations of 3DGS in modeling highly specular surfaces using Spherical Harmonics (SH). Third, as observed in our ablation studies, while the full combined model achieves the highest average PSNR, individual variants slightly outperform it on SSIM and LPIPS. This reveals that a consistently optimal optimization method that uniformly maximizes all evaluation metrics has not been found.

Future work will focus on reducing the CDF computational overhead and exploring more comprehensive optimization algorithms to jointly improve both reconstruction fidelity and perceptual quality. Furthermore, rather than applying complex asymmetric kernels globally, we plan to integrate SNS with spatial hierarchical structures. This would allow the model to adaptively allocate Skew-Normal primitives along high-frequency object boundaries while falling back to simpler Gaussians in smooth regions, thereby optimizing the overall efficiency-to-quality trade-off.

\end{document}

%% file: figures/v3.pdf_tex

\begingroup%
\makeatletter%
\providecommand\color[2][]{%
  \errmessage{(Inkscape) Color is used for the text in Inkscape, but the package 'color.sty' is not loaded}%
  \renewcommand\color[2][]{}%
}%
\providecommand\transparent[1]{%
  \errmessage{(Inkscape) Transparency is used (non-zero) for the text in Inkscape, but the package 'transparent.sty' is not loaded}%
  \renewcommand\transparent[1]{}%
}%
\providecommand\rotatebox[2]{#2}%
\newcommand*\fsize{\dimexpr\f@size pt\relax}%
\newcommand*\lineheight[1]{\fontsize{\fsize}{#1\fsize}\selectfont}%
\ifx\svgwidth\undefined%
  \setlength{\unitlength}{412.14959092bp}%
  \ifx\svgscale\undefined%
    \relax%
  \else%
    \setlength{\unitlength}{\unitlength * \real{\svgscale}}%
  \fi%
\else%
  \setlength{\unitlength}{\svgwidth}%
\fi%
\global\let\svgwidth\undefined%
\global\let\svgscale\undefined%
\makeatother%
\begin{picture}(1,0.90347676)%
  \lineheight{1}%
  \setlength\tabcolsep{0pt}%

  \put(0.07859362,-0.02588231){\color[rgb]{0,0,0}\makebox(0,0)[t]{\lineheight{1.25}\smash{\begin{tabular}[t]{c}(a) kitchen\end{tabular}}}}%
  \put(0.23447435,-0.02588231){\color[rgb]{0,0,0}\makebox(0,0)[lt]{\lineheight{1.25}\smash{\begin{tabular}[t]{l}(b) bonsai\end{tabular}}}}%
  \put(0.43743576,-0.02588231){\color[rgb]{0,0,0}\makebox(0,0)[lt]{\lineheight{1.25}\smash{\begin{tabular}[t]{l}(c) room\end{tabular}}}}%
  \put(0.65365327,-0.02588231){\color[rgb]{0,0,0}\makebox(0,0)[lt]{\lineheight{1.25}\smash{\begin{tabular}[t]{l}(d) train\end{tabular}}}}%
  \put(0.84599965,-0.0257867){\color[rgb]{0,0,0}\makebox(0,0)[lt]{\lineheight{1.25}\smash{\begin{tabular}[t]{l}(e) drjohnson\end{tabular}}}}%

  \put(-0.01242011,0.04650598){\color[rgb]{0,0,0}\rotatebox{90}{\makebox(0,0)[lt]{\lineheight{1.25}\smash{\footnotesize\begin{tabular}[t]{l}Ours\end{tabular}}}}}%
  \put(-0.01242011,0.17948882){\color[rgb]{0,0,0}\rotatebox{90}{\makebox(0,0)[lt]{\lineheight{1.25}\smash{\footnotesize\begin{tabular}[t]{l}SSS\end{tabular}}}}}%
  \put(-0.01243696,0.26746146){\color[rgb]{0,0,0}\rotatebox{90}{\makebox(0,0)[lt]{\lineheight{1.25}\smash{\footnotesize\begin{tabular}[t]{l}3DGS-MCMC\end{tabular}}}}}%
  \put(-0.01242011,0.6997214){\color[rgb]{0,0,0}\rotatebox{90}{\makebox(0,0)[lt]{\lineheight{1.25}\smash{\footnotesize\begin{tabular}[t]{l}3DGS\end{tabular}}}}}%
  \put(-0.01242011,0.56223443){\color[rgb]{0,0,0}\rotatebox{90}{\makebox(0,0)[lt]{\lineheight{1.25}\smash{\footnotesize\begin{tabular}[t]{l}GES\end{tabular}}}}}%
  \put(-0.01245381,0.41934001){\color[rgb]{0,0,0}\rotatebox{90}{\makebox(0,0)[lt]{\lineheight{1.25}\smash{\footnotesize\begin{tabular}[t]{l}3DHGS\end{tabular}}}}}%
  \put(-0.01242009,0.82594888){\color[rgb]{0,0,0}\rotatebox{90}{\makebox(0,0)[lt]{\lineheight{1.25}\smash{\footnotesize\begin{tabular}[t]{l}GT\end{tabular}}}}}%

  \put(0,0){\includegraphics[width=\unitlength,page=1]{figures/v3.pdf}}%
\end{picture}%
\endgroup%

%% file: main.bib
@inproceedings{DBLP:conf/eccv/YaoLLFQ18,
  author       = {Yao Yao and
                  Zixin Luo and
                  Shiwei Li and
                  Tian Fang and
                  Long Quan},
  editor       = {Vittorio Ferrari and
                  Martial Hebert and
                  Cristian Sminchisescu and
                  Yair Weiss},
  title        = {{{MVSNet}: Depth Inference for Unstructured Multi-view Stereo}},
  booktitle    = {Computer Vision - {ECCV} 2018 - 15th European Conference, Munich,
                  Germany, September 8-14, 2018, Proceedings, Part {VIII}},
  series       = {Lecture Notes in Computer Science},
  pages        = {785--801},
  publisher    = {Springer},
  year         = {2018},
  url          = {https://doi.org/10.1007/978-3-030-01237-3\_47},
  doi          = {10.1007/978-3-030-01237-3\_47},
  bibsource    = {dblp computer science bibliography, https://dblp.org}
}

@inproceedings{DBLP:conf/cvpr/SchonbergerF16,
  author       = {Johannes L. Sch{\"{o}}nberger and
                  Jan{-}Michael Frahm},
  title        = {{Structure-from-Motion Revisited}},
  booktitle    = {2016 {IEEE} Conference on Computer Vision and Pattern Recognition,
                  {CVPR} 2016, Las Vegas, NV, USA, June 27-30, 2016},
  pages        = {4104--4113},
  publisher    = {{IEEE} Computer Society},
  year         = {2016},
  url          = {https://doi.org/10.1109/CVPR.2016.445},
  doi          = {10.1109/CVPR.2016.445},
  bibsource    = {dblp computer science bibliography, https://dblp.org}
}

@inproceedings{DBLP:conf/eccv/WeiZLFX20,
  author       = {Xingkui Wei and
                  Yinda Zhang and
                  Zhuwen Li and
                  Yanwei Fu and
                  Xiangyang Xue},
  editor       = {Andrea Vedaldi and
                  Horst Bischof and
                  Thomas Brox and
                  Jan{-}Michael Frahm},
  title        = {{{DeepSFM}: Structure from Motion via Deep Bundle Adjustment}},
  booktitle    = {Computer Vision - {ECCV} 2020 - 16th European Conference, Glasgow,
                  UK, August 23-28, 2020, Proceedings, Part {I}},
  series       = {Lecture Notes in Computer Science},
  pages        = {230--247},
  publisher    = {Springer},
  year         = {2020},
  url          = {https://doi.org/10.1007/978-3-030-58452-8\_14},
  doi          = {10.1007/978-3-030-58452-8\_14},
  bibsource    = {dblp computer science bibliography, https://dblp.org}
}

@article{DBLP:journals/tmlr/CaoRF22,
  author       = {Chenjie Cao and
                  Xinlin Ren and
                  Yanwei Fu},
  title        = {{{MVSFormer}: Multi-View Stereo by Learning Robust Image Features and
                  Temperature-based Depth}},
  journal      = {Trans. Mach. Learn. Res.},
  volume       = {2022},
  year         = {2022},
  url          = {https://openreview.net/forum?id=2VWR6JfwNo},
  bibsource    = {dblp computer science bibliography, https://dblp.org}
}

@article{DBLP:journals/cacm/MildenhallSTBRN22,
  author       = {Ben Mildenhall and
                  Pratul P. Srinivasan and
                  Matthew Tancik and
                  Jonathan T. Barron and
                  Ravi Ramamoorthi and
                  Ren Ng},
  title        = {{{NeRF}: representing scenes as neural radiance fields for view synthesis}},
  journal      = {Commun. {ACM}},
  volume       = {65},
  number       = {1},
  pages        = {99--106},
  year         = {2022},
  url          = {https://doi.org/10.1145/3503250},
  doi          = {10.1145/3503250},
  bibsource    = {dblp computer science bibliography, https://dblp.org}
}

@inproceedings{DBLP:conf/eccv/MildenhallSTBRN20,
  author       = {Ben Mildenhall and
                  Pratul P. Srinivasan and
                  Matthew Tancik and
                  Jonathan T. Barron and
                  Ravi Ramamoorthi and
                  Ren Ng},
  editor       = {Andrea Vedaldi and
                  Horst Bischof and
                  Thomas Brox and
                  Jan{-}Michael Frahm},
  title        = {{{NeRF}: Representing Scenes as Neural Radiance Fields for View Synthesis}},
  booktitle    = {Computer Vision - {ECCV} 2020 - 16th European Conference, Glasgow,
                  UK, August 23-28, 2020, Proceedings, Part {I}},
  series       = {Lecture Notes in Computer Science},
  pages        = {405--421},
  publisher    = {Springer},
  year         = {2020},
  url          = {https://doi.org/10.1007/978-3-030-58452-8\_24},
  doi          = {10.1007/978-3-030-58452-8\_24},
  bibsource    = {dblp computer science bibliography, https://dblp.org}
}

@inproceedings{DBLP:conf/iccv/BarronMTHMS21,
  author       = {Jonathan T. Barron and
                  Ben Mildenhall and
                  Matthew Tancik and
                  Peter Hedman and
                  Ricardo Martin{-}Brualla and
                  Pratul P. Srinivasan},
  title        = {{Mip-{NeRF}: {A} Multiscale Representation for Anti-Aliasing Neural Radiance
                  Fields}},
  booktitle    = {2021 {IEEE/CVF} International Conference on Computer Vision, {ICCV}
                  2021, Montreal, QC, Canada, October 10-17, 2021},
  pages        = {5835--5844},
  publisher    = {{IEEE}},
  year         = {2021},
  url          = {https://doi.org/10.1109/ICCV48922.2021.00580},
  doi          = {10.1109/ICCV48922.2021.00580},
  bibsource    = {dblp computer science bibliography, https://dblp.org}
}

@inproceedings{DBLP:conf/cvpr/BarronMVSH22,
  author       = {Jonathan T. Barron and
                  Ben Mildenhall and
                  Dor Verbin and
                  Pratul P. Srinivasan and
                  Peter Hedman},
  title        = {{Mip-{NeRF} 360: Unbounded Anti-Aliased Neural Radiance Fields}},
  booktitle    = {{IEEE/CVF} Conference on Computer Vision and Pattern Recognition,
                  {CVPR} 2022, New Orleans, LA, USA, June 18-24, 2022},
  pages        = {5460--5469},
  publisher    = {{IEEE}},
  year         = {2022},
  url          = {https://doi.org/10.1109/CVPR52688.2022.00539},
  doi          = {10.1109/CVPR52688.2022.00539},
  bibsource    = {dblp computer science bibliography, https://dblp.org}
}

@inproceedings{DBLP:conf/cvpr/Fridovich-KeilY22,
  author       = {Sara Fridovich{-}Keil and
                  Alex Yu and
                  Matthew Tancik and
                  Qinhong Chen and
                  Benjamin Recht and
                  Angjoo Kanazawa},
  title        = {{Plenoxels: Radiance Fields without Neural Networks}},
  booktitle    = {{IEEE/CVF} Conference on Computer Vision and Pattern Recognition,
                  {CVPR} 2022, New Orleans, LA, USA, June 18-24, 2022},
  pages        = {5491--5500},
  publisher    = {{IEEE}},
  year         = {2022},
  url          = {https://doi.org/10.1109/CVPR52688.2022.00542},
  doi          = {10.1109/CVPR52688.2022.00542},
  bibsource    = {dblp computer science bibliography, https://dblp.org}
}

@inproceedings{DBLP:conf/cvpr/Fridovich-KeilM23,
  author       = {Sara Fridovich{-}Keil and
                  Giacomo Meanti and
                  Frederik Rahb{\ae}k Warburg and
                  Benjamin Recht and
                  Angjoo Kanazawa},
  title        = {{{K-Planes}: Explicit Radiance Fields in Space, Time, and Appearance}},
  booktitle    = {{IEEE/CVF} Conference on Computer Vision and Pattern Recognition,
                  {CVPR} 2023, Vancouver, BC, Canada, June 17-24, 2023},
  pages        = {12479--12488},
  publisher    = {{IEEE}},
  year         = {2023},
  url          = {https://doi.org/10.1109/CVPR52729.2023.01201},
  doi          = {10.1109/CVPR52729.2023.01201},
  bibsource    = {dblp computer science bibliography, https://dblp.org}
}

@inproceedings{DBLP:conf/cvpr/YuYTK21,
  author       = {Alex Yu and
                  Vickie Ye and
                  Matthew Tancik and
                  Angjoo Kanazawa},
  title        = {{pixel{NeRF}: Neural Radiance Fields From One or Few Images}},
  booktitle    = {{IEEE} Conference on Computer Vision and Pattern Recognition, {CVPR}
                  2021, virtual, June 19-25, 2021},
  pages        = {4578--4587},
  publisher    = {Computer Vision Foundation / {IEEE}},
  year         = {2021},
  url          = {https://openaccess.thecvf.com/content/CVPR2021/html/Yu\_pixel{NeRF}\_Neural\_Radiance\_Fields\_From\_One\_or\_Few\_Images\_CVPR\_2021\_paper.html},
  doi          = {10.1109/CVPR46437.2021.00455},
  bibsource    = {dblp computer science bibliography, https://dblp.org}
}

@inproceedings{DBLP:conf/cvpr/WangWGSZBMSF21,
  author       = {Qianqian Wang and
                  Zhicheng Wang and
                  Kyle Genova and
                  Pratul P. Srinivasan and
                  Howard Zhou and
                  Jonathan T. Barron and
                  Ricardo Martin{-}Brualla and
                  Noah Snavely and
                  Thomas A. Funkhouser},
  title        = {{{IBRNet}: Learning Multi-View Image-Based Rendering}},
  booktitle    = {{IEEE} Conference on Computer Vision and Pattern Recognition, {CVPR}
                  2021, virtual, June 19-25, 2021},
  pages        = {4690--4699},
  publisher    = {Computer Vision Foundation / {IEEE}},
  year         = {2021},
  url          = {https://openaccess.thecvf.com/content/CVPR2021/html/Wang\_{IBRNet}\_Learning\_Multi-View\_Image-Based\_Rendering\_CVPR\_2021\_paper.html},
  doi          = {10.1109/CVPR46437.2021.00466},
  bibsource    = {dblp computer science bibliography, https://dblp.org}
}

@inproceedings{DBLP:conf/iclr/YuG022,
  author       = {Hong{-}Xing Yu and
                  Leonidas J. Guibas and
                  Jiajun Wu},
  title        = {{Unsupervised Discovery of Object Radiance Fields}},
  booktitle    = {The Tenth International Conference on Learning Representations, {ICLR}
                  2022, Virtual Event, April 25-29, 2022},
  publisher    = {OpenReview.net},
  year         = {2022},
  url          = {https://openreview.net/forum?id=rwE8SshAlxw},
  bibsource    = {dblp computer science bibliography, https://dblp.org}
}

@inproceedings{DBLP:conf/nips/QiYZ23,
  author       = {Di Qi and
                  Tong Yang and
                  Xiangyu Zhang},
  editor       = {Alice Oh and
                  Tristan Naumann and
                  Amir Globerson and
                  Kate Saenko and
                  Moritz Hardt and
                  Sergey Levine},
  title        = {{Slot-guided Volumetric Object Radiance Fields}},
  booktitle    = {Advances in Neural Information Processing Systems 36: Annual Conference
                  on Neural Information Processing Systems 2023, NeurIPS 2023, New Orleans,
                  LA, USA, December 10 - 16, 2023},
  year         = {2023},
  url          = {http://papers.nips.cc/paper\_files/paper/2023/hash/cf66f995883298c4db2f0dcba28fb211-Abstract-Conference.html},
  bibsource    = {dblp computer science bibliography, https://dblp.org}
}

@inproceedings{DBLP:conf/nips/YarivGKL21,
  author       = {Lior Yariv and
                  Jiatao Gu and
                  Yoni Kasten and
                  Yaron Lipman},
  editor       = {Marc'Aurelio Ranzato and
                  Alina Beygelzimer and
                  Yann N. Dauphin and
                  Percy Liang and
                  Jennifer Wortman Vaughan},
  title        = {{Volume Rendering of Neural Implicit Surfaces}},
  booktitle    = {Advances in Neural Information Processing Systems 34: Annual Conference
                  on Neural Information Processing Systems 2021, NeurIPS 2021, December
                  6-14, 2021, virtual},
  pages        = {4805--4815},
  year         = {2021},
  url          = {https://proceedings.neurips.cc/paper/2021/hash/25e2a30f44898b9f3e978b1786dcd85c-Abstract.html},
  bibsource    = {dblp computer science bibliography, https://dblp.org}
}

@inproceedings{DBLP:conf/nips/WangLLTKW21,
  author       = {Peng Wang and
                  Lingjie Liu and
                  Yuan Liu and
                  Christian Theobalt and
                  Taku Komura and
                  Wenping Wang},
  editor       = {Marc'Aurelio Ranzato and
                  Alina Beygelzimer and
                  Yann N. Dauphin and
                  Percy Liang and
                  Jennifer Wortman Vaughan},
  title        = {{{NeuS}: Learning Neural Implicit Surfaces by Volume Rendering for Multi-view
                  Reconstruction}},
  booktitle    = {Advances in Neural Information Processing Systems 34: Annual Conference
                  on Neural Information Processing Systems 2021, NeurIPS 2021, December
                  6-14, 2021, virtual},
  pages        = {27171--27183},
  year         = {2021},
  url          = {https://proceedings.neurips.cc/paper/2021/hash/e41e164f7485ec4a28741a2d0ea41c74-Abstract.html},
  bibsource    = {dblp computer science bibliography, https://dblp.org}
}

@inproceedings{DBLP:conf/cvpr/PumarolaCPM21,
  author       = {Albert Pumarola and
                  Enric Corona and
                  Gerard Pons{-}Moll and
                  Francesc Moreno{-}Noguer},
  title        = {{D-{NeRF}: Neural Radiance Fields for Dynamic Scenes}},
  booktitle    = {{IEEE} Conference on Computer Vision and Pattern Recognition, {CVPR}
                  2021, virtual, June 19-25, 2021},
  pages        = {10318--10327},
  publisher    = {Computer Vision Foundation / {IEEE}},
  year         = {2021},
  url          = {https://openaccess.thecvf.com/content/CVPR2021/html/Pumarola\_D-{NeRF}\_Neural\_Radiance\_Fields\_for\_Dynamic\_Scenes\_CVPR\_2021\_paper.html},
  doi          = {10.1109/CVPR46437.2021.01018},
  bibsource    = {dblp computer science bibliography, https://dblp.org}
}

@article{DBLP:journals/tog/ParkSHBBGMS21,
  author       = {Keunhong Park and
                  Utkarsh Sinha and
                  Peter Hedman and
                  Jonathan T. Barron and
                  Sofien Bouaziz and
                  Dan B. Goldman and
                  Ricardo Martin{-}Brualla and
                  Steven M. Seitz},
  title        = {{Hyper{NeRF}: a higher-dimensional representation for topologically varying
                  neural radiance fields}},
  journal      = {{ACM} Trans. Graph.},
  volume       = {40},
  number       = {6},
  pages        = {238:1--238:12},
  year         = {2021},
  url          = {https://doi.org/10.1145/3478513.3480487},
  doi          = {10.1145/3478513.3480487},
  bibsource    = {dblp computer science bibliography, https://dblp.org}
}

@article{DBLP:journals/tog/KerblKLD23,
  author       = {Bernhard Kerbl and
                  Georgios Kopanas and
                  Thomas Leimk{\"{u}}hler and
                  George Drettakis},
  title        = {{{3D} {Gaussian Splatting} for Real-Time Radiance Field Rendering}},
  journal      = {{ACM} Trans. Graph.},
  volume       = {42},
  number       = {4},
  pages        = {139:1--139:14},
  year         = {2023},
  url          = {https://doi.org/10.1145/3592433},
  doi          = {10.1145/3592433},
  bibsource    = {dblp computer science bibliography, https://dblp.org}
}

@inproceedings{DBLP:conf/cvpr/YuCHS024,
  author       = {Zehao Yu and
                  Anpei Chen and
                  Binbin Huang and
                  Torsten Sattler and
                  Andreas Geiger},
  title        = {{Mip-{Splatting}: Alias-Free {3D} {Gaussian Splatting}}},
  booktitle    = {{IEEE/CVF} Conference on Computer Vision and Pattern Recognition,
                  {CVPR} 2024, Seattle, WA, USA, June 16-22, 2024},
  pages        = {19447--19456},
  publisher    = {{IEEE}},
  year         = {2024},
  url          = {https://doi.org/10.1109/CVPR52733.2024.01839},
  doi          = {10.1109/CVPR52733.2024.01839},
  bibsource    = {dblp computer science bibliography, https://dblp.org}
}

@inproceedings{DBLP:conf/cvpr/0005YXX0L024,
  author       = {Tao Lu and
                  Mulin Yu and
                  Linning Xu and
                  Yuanbo Xiangli and
                  Limin Wang and
                  Dahua Lin and
                  Bo Dai},
  title        = {{{Scaffold-GS}: Structured {3D} Gaussians for View-Adaptive Rendering}},
  booktitle    = {{IEEE/CVF} Conference on Computer Vision and Pattern Recognition,
                  {CVPR} 2024, Seattle, WA, USA, June 16-22, 2024},
  pages        = {20654--20664},
  publisher    = {{IEEE}},
  year         = {2024},
  url          = {https://doi.org/10.1109/CVPR52733.2024.01952},
  doi          = {10.1109/CVPR52733.2024.01952},
  bibsource    = {dblp computer science bibliography, https://dblp.org}
}

@ARTICLE{10993308,
  author={Ren, Kerui and Jiang, Lihan and Lu, Tao and Yu, Mulin and Xu, Linning and Ni, Zhangkai and Dai, Bo},
  journal={IEEE Transactions on Pattern Analysis and Machine Intelligence},
  title={{{Octree-GS}: Towards Consistent Real-time Rendering with LOD-Structured {3D} Gaussians}},
  year={2025},
  volume={},
  number={},
  pages={1-15},
  doi={10.1109/TPAMI.2025.3568201}}

@inproceedings{DBLP:conf/eccv/ZhangHLHZ24,
  author       = {Zheng Zhang and
                  Wenbo Hu and
                  Yixing Lao and
                  Tong He and
                  Hengshuang Zhao},
  editor       = {Ales Leonardis and
                  Elisa Ricci and
                  Stefan Roth and
                  Olga Russakovsky and
                  Torsten Sattler and
                  G{\"{u}}l Varol},
  title        = {{Pixel-{GS}: Density Control with Pixel-Aware Gradient for {3D} Gaussian
                  Splatting}},
  booktitle    = {Computer Vision - {ECCV} 2024 - 18th European Conference, Milan, Italy,
                  September 29-October 4, 2024, Proceedings, Part {XIX}},
  series       = {Lecture Notes in Computer Science},
  pages        = {326--342},
  publisher    = {Springer},
  year         = {2024},
  url          = {https://doi.org/10.1007/978-3-031-72655-2\_19},
  doi          = {10.1007/978-3-031-72655-2\_19},
  bibsource    = {dblp computer science bibliography, https://dblp.org}
}

@inproceedings{10.1145/3664647.3681361,
author = {Ye, Zongxin and Li, Wenyu and Liu, Sidun and Qiao, Peng and Dou, Yong},
title = {{{AbsGS}: Recovering Fine Details in {3D} {Gaussian Splatting}}},
year = {2024},
isbn = {9798400706868},
publisher = {Association for Computing Machinery},
address = {New York, NY, USA},
url = {https://doi.org/10.1145/3664647.3681361},
doi = {10.1145/3664647.3681361},
booktitle = {Proceedings of the 32nd ACM International Conference on Multimedia},
pages = {1053–1061},
numpages = {9},
location = {Melbourne VIC, Australia},
series = {MM '24}
}

@inproceedings{DBLP:conf/cvpr/WangF0X024,
  author       = {Junjie Wang and
                  Jiemin Fang and
                  Xiaopeng Zhang and
                  Lingxi Xie and
                  Qi Tian},
  title        = {{{GaussianEditor}: Editing {3D} Gaussians Delicately with Text Instructions}},
  booktitle    = {{IEEE/CVF} Conference on Computer Vision and Pattern Recognition,
                  {CVPR} 2024, Seattle, WA, USA, June 16-22, 2024},
  pages        = {20902--20911},
  publisher    = {{IEEE}},
  year         = {2024},
  url          = {https://doi.org/10.1109/CVPR52733.2024.01975},
  doi          = {10.1109/CVPR52733.2024.01975},
  bibsource    = {dblp computer science bibliography, https://dblp.org}
}

@inproceedings{DBLP:conf/cvpr/ChenCZWYWCYLL24,
  author       = {Yiwen Chen and
                  Zilong Chen and
                  Chi Zhang and
                  Feng Wang and
                  Xiaofeng Yang and
                  Yikai Wang and
                  Zhongang Cai and
                  Lei Yang and
                  Huaping Liu and
                  Guosheng Lin},
  title        = {{{GaussianEditor}: Swift and Controllable {3D} Editing with {Gaussian Splatting}}},
  booktitle    = {{IEEE/CVF} Conference on Computer Vision and Pattern Recognition,
                  {CVPR} 2024, Seattle, WA, USA, June 16-22, 2024},
  pages        = {21476--21485},
  publisher    = {{IEEE}},
  year         = {2024},
  url          = {https://doi.org/10.1109/CVPR52733.2024.02029},
  doi          = {10.1109/CVPR52733.2024.02029},
  bibsource    = {dblp computer science bibliography, https://dblp.org}
}

@inproceedings{DBLP:conf/cvpr/WuYFX0000W24,
  author       = {Guanjun Wu and
                  Taoran Yi and
                  Jiemin Fang and
                  Lingxi Xie and
                  Xiaopeng Zhang and
                  Wei Wei and
                  Wenyu Liu and
                  Qi Tian and
                  Xinggang Wang},
  title        = {{{4D {Gaussian Splatting}} for Real-Time Dynamic Scene Rendering}},
  booktitle    = {{IEEE/CVF} Conference on Computer Vision and Pattern Recognition,
                  {CVPR} 2024, Seattle, WA, USA, June 16-22, 2024},
  pages        = {20310--20320},
  publisher    = {{IEEE}},
  year         = {2024},
  url          = {https://doi.org/10.1109/CVPR52733.2024.01920},
  doi          = {10.1109/CVPR52733.2024.01920},
  bibsource    = {dblp computer science bibliography, https://dblp.org}
}

@inproceedings{DBLP:conf/iclr/YangYP024,
  author       = {Zeyu Yang and
                  Hongye Yang and
                  Zijie Pan and
                  Li Zhang},
  title        = {{Real-time Photorealistic Dynamic Scene Representation and Rendering
                  with {4D {Gaussian Splatting}}}},
  booktitle    = {The Twelfth International Conference on Learning Representations,
                  {ICLR} 2024, Vienna, Austria, May 7-11, 2024},
  publisher    = {OpenReview.net},
  year         = {2024},
  url          = {https://openreview.net/forum?id=WhgB5sispV},
  bibsource    = {dblp computer science bibliography, https://dblp.org}
}

@inproceedings{DBLP:conf/cvpr/YiFWWX000W24,
  author       = {Taoran Yi and
                  Jiemin Fang and
                  Junjie Wang and
                  Guanjun Wu and
                  Lingxi Xie and
                  Xiaopeng Zhang and
                  Wenyu Liu and
                  Qi Tian and
                  Xinggang Wang},
  title        = {{{GaussianDreamer}: Fast Generation from Text to {3D} Gaussians by Bridging
                  2D and {3D} Diffusion Models}},
  booktitle    = {{IEEE/CVF} Conference on Computer Vision and Pattern Recognition,
                  {CVPR} 2024, Seattle, WA, USA, June 16-22, 2024},
  pages        = {6796--6807},
  publisher    = {{IEEE}},
  year         = {2024},
  url          = {https://doi.org/10.1109/CVPR52733.2024.00649},
  doi          = {10.1109/CVPR52733.2024.00649},
  bibsource    = {dblp computer science bibliography, https://dblp.org}
}

@inproceedings{DBLP:conf/iclr/TangRZ0Z24,
  author       = {Jiaxiang Tang and
                  Jiawei Ren and
                  Hang Zhou and
                  Ziwei Liu and
                  Gang Zeng},
  title        = {{{DreamGaussian}: Generative {Gaussian Splatting} for Efficient {3D} Content
                  Creation}},
  booktitle    = {The Twelfth International Conference on Learning Representations,
                  {ICLR} 2024, Vienna, Austria, May 7-11, 2024},
  publisher    = {OpenReview.net},
  year         = {2024},
  url          = {https://openreview.net/forum?id=UyNXMqnN3c},
  bibsource    = {dblp computer science bibliography, https://dblp.org}
}

@inproceedings{DBLP:conf/eccv/XuSWCYPSW24,
  author       = {Yinghao Xu and
                  Zifan Shi and
                  Yifan Wang and
                  Hansheng Chen and
                  Ceyuan Yang and
                  Sida Peng and
                  Yujun Shen and
                  Gordon Wetzstein},
  editor       = {Ales Leonardis and
                  Elisa Ricci and
                  Stefan Roth and
                  Olga Russakovsky and
                  Torsten Sattler and
                  G{\"{u}}l Varol},
  title        = {{{GRM:} Large Gaussian Reconstruction Model for Efficient {3D} Reconstruction
                  and Generation}},
  booktitle    = {Computer Vision - {ECCV} 2024 - 18th European Conference, Milan, Italy,
                  September 29-October 4, 2024, Proceedings, Part {XV}},
  series       = {Lecture Notes in Computer Science},
  pages        = {1--20},
  publisher    = {Springer},
  year         = {2024},
  url          = {https://doi.org/10.1007/978-3-031-72633-0\_1},
  doi          = {10.1007/978-3-031-72633-0\_1},
  bibsource    = {dblp computer science bibliography, https://dblp.org}
}

@inproceedings{DBLP:conf/eccv/TangCCWZL24,
  author       = {Jiaxiang Tang and
                  Zhaoxi Chen and
                  Xiaokang Chen and
                  Tengfei Wang and
                  Gang Zeng and
                  Ziwei Liu},
  editor       = {Ales Leonardis and
                  Elisa Ricci and
                  Stefan Roth and
                  Olga Russakovsky and
                  Torsten Sattler and
                  G{\"{u}}l Varol},
  title        = {{{LGM:} Large Multi-view Gaussian Model for High-Resolution {3D} Content
                  Creation}},
  booktitle    = {Computer Vision - {ECCV} 2024 - 18th European Conference, Milan, Italy,
                  September 29-October 4, 2024, Proceedings, Part {IV}},
  series       = {Lecture Notes in Computer Science},
  pages        = {1--18},
  publisher    = {Springer},
  year         = {2024},
  url          = {https://doi.org/10.1007/978-3-031-73235-5\_1},
  doi          = {10.1007/978-3-031-73235-5\_1},
  bibsource    = {dblp computer science bibliography, https://dblp.org}
}

@inproceedings{DBLP:conf/cvpr/Qin0ZWP24,
  author       = {Minghan Qin and
                  Wanhua Li and
                  Jiawei Zhou and
                  Haoqian Wang and
                  Hanspeter Pfister},
  title        = {{{LangSplat}: {3D} Language {Gaussian Splatting}}},
  booktitle    = {{IEEE/CVF} Conference on Computer Vision and Pattern Recognition,
                  {CVPR} 2024, Seattle, WA, USA, June 16-22, 2024},
  pages        = {20051--20060},
  publisher    = {{IEEE}},
  year         = {2024},
  url          = {https://doi.org/10.1109/CVPR52733.2024.01895},
  doi          = {10.1109/CVPR52733.2024.01895},
  bibsource    = {dblp computer science bibliography, https://dblp.org}
}

@inproceedings{DBLP:conf/nips/WuMLWSC0FD0024,
  author       = {Yanmin Wu and
                  Jiarui Meng and
                  Haijie Li and
                  Chenming Wu and
                  Yahao Shi and
                  Xinhua Cheng and
                  Chen Zhao and
                  Haocheng Feng and
                  Errui Ding and
                  Jingdong Wang and
                  Jian Zhang},
  editor       = {Amir Globersons and
                  Lester Mackey and
                  Danielle Belgrave and
                  Angela Fan and
                  Ulrich Paquet and
                  Jakub M. Tomczak and
                  Cheng Zhang},
  title        = {{{OpenGaussian}: Towards Point-Level {3D} Gaussian-based Open Vocabulary
                  Understanding}},
  booktitle    = {Advances in Neural Information Processing Systems 38: Annual Conference
                  on Neural Information Processing Systems 2024, NeurIPS 2024, Vancouver,
                  BC, Canada, December 10 - 15, 2024},
  year         = {2024},
  url          = {http://papers.nips.cc/paper\_files/paper/2024/hash/21f7b745f73ce0d1f9bcea7f40b1388e-Abstract-Conference.html},
  bibsource    = {dblp computer science bibliography, https://dblp.org}
}

@inproceedings{DBLP:conf/eccv/YeDYK24,
  author       = {Mingqiao Ye and
                  Martin Danelljan and
                  Fisher Yu and
                  Lei Ke},
  editor       = {Ales Leonardis and
                  Elisa Ricci and
                  Stefan Roth and
                  Olga Russakovsky and
                  Torsten Sattler and
                  G{\"{u}}l Varol},
  title        = {{Gaussian Grouping: Segment and Edit Anything in {3D} Scenes}},
  booktitle    = {Computer Vision - {ECCV} 2024 - 18th European Conference, Milan, Italy,
                  September 29-October 4, 2024, Proceedings, Part {XXIX}},
  series       = {Lecture Notes in Computer Science},
  pages        = {162--179},
  publisher    = {Springer},
  year         = {2024},
  url          = {https://doi.org/10.1007/978-3-031-73397-0\_10},
  doi          = {10.1007/978-3-031-73397-0\_10},
  bibsource    = {dblp computer science bibliography, https://dblp.org}
}

@inproceedings{DBLP:conf/aaai/CenF0X00025,
  author       = {Jiazhong Cen and
                  Jiemin Fang and
                  Chen Yang and
                  Lingxi Xie and
                  Xiaopeng Zhang and
                  Wei Shen and
                  Qi Tian},
  editor       = {Toby Walsh and
                  Julie Shah and
                  Zico Kolter},
  title        = {{Segment Any {3D} Gaussians}},
  booktitle    = {Thirty-Ninth {AAAI} Conference on Artificial Intelligence, Thirty-Seventh
                  Conference on Innovative Applications of Artificial Intelligence,
                  Fifteenth Symposium on Educational Advances in Artificial Intelligence,
                  {AAAI} 2025, Philadelphia, PA, USA, February 25 - March 4, 2025},
  pages        = {1971--1979},
  publisher    = {{AAAI} Press},
  year         = {2025},
  url          = {https://doi.org/10.1609/aaai.v39i2.32193},
  doi          = {10.1609/AAAI.V39I2.32193},
  bibsource    = {dblp computer science bibliography, https://dblp.org}
}

@inproceedings{DBLP:conf/cvpr/HamdiMMQLVGV24,
  author       = {Abdullah Hamdi and
                  Luke Melas{-}Kyriazi and
                  Jinjie Mai and
                  Guocheng Qian and
                  Ruoshi Liu and
                  Carl Vondrick and
                  Bernard Ghanem and
                  Andrea Vedaldi},
  title        = {{{GES:} Generalized Exponential Splatting for Efficient Radiance Field
                  Rendering}},
  booktitle    = {{IEEE/CVF} Conference on Computer Vision and Pattern Recognition,
                  {CVPR} 2024, Seattle, WA, USA, June 16-22, 2024},
  pages        = {19812--19822},
  publisher    = {{IEEE}},
  year         = {2024},
  url          = {https://doi.org/10.1109/CVPR52733.2024.01873},
  doi          = {10.1109/CVPR52733.2024.01873},
  bibsource    = {dblp computer science bibliography, https://dblp.org}
}

@inproceedings{DBLP:conf/cvpr/ZhuYH025,
  author       = {Jialin Zhu and
                  Jiangbei Yue and
                  Feixiang He and
                  He Wang},
  title        = {{{3D} Student Splatting and Scooping}},
  booktitle    = {{IEEE/CVF} Conference on Computer Vision and Pattern Recognition,
                  {CVPR} 2025, Nashville, TN, USA, June 11-15, 2025},
  pages        = {21045--21054},
  publisher    = {Computer Vision Foundation / {IEEE}},
  year         = {2025},
  url          = {https://openaccess.thecvf.com/content/CVPR2025/html/Zhu\_{3D}\_Student\_Splatting\_and\_Scooping\_CVPR\_2025\_paper.html},
  doi          = {10.1109/CVPR52734.2025.01960},
  bibsource    = {dblp computer science bibliography, https://dblp.org}
}

@inproceedings{DBLP:conf/cvpr/Li0SC25,
  author       = {Haolin Li and
                  Jinyang Liu and
                  Mario Sznaier and
                  Octavia I. Camps},
  title        = {{{{3D}-HGS}: {3D} Half-{Gaussian Splatting}}},
  booktitle    = {{IEEE/CVF} Conference on Computer Vision and Pattern Recognition,
                  {CVPR} 2025, Nashville, TN, USA, June 11-15, 2025},
  pages        = {10996--11005},
  publisher    = {Computer Vision Foundation / {IEEE}},
  year         = {2025},
  url          = {https://openaccess.thecvf.com/content/CVPR2025/html/Li\_{{3D}-HGS}\_{3D}\_Half-Gaussian\_Splatting\_CVPR\_2025\_paper.html},
  doi          = {10.1109/CVPR52734.2025.01027},
  bibsource    = {dblp computer science bibliography, https://dblp.org}
}

@ARTICLE{11442667,
  author={Guo, Jun and Ma, Xiaojian and Fan, Yue and Liu, Huaping and Li, Qing},
  journal={IEEE Transactions on Circuits and Systems for Video Technology},
  title={{Semantic Gaussians: Open-Vocabulary Scene Understanding with {3D} {Gaussian Splatting}}},
  year={2026},
  volume={},
  number={},
  pages={1-1},
  doi={10.1109/TCSVT.2026.3675320}}

@article{Azzalini_1999,
   title={{Statistical Applications of the Multivariate Skew Normal Distribution}},
   volume={61},
   ISSN={1467-9868},
   url={http://dx.doi.org/10.1111/1467-9868.00194},
   DOI={10.1111/1467-9868.00194},
   number={3},
   journal={Journal of the Royal Statistical Society Series B: Statistical Methodology},
   publisher={Oxford University Press (OUP)},
   author={Azzalini, A. and Capitanio, A.},
   year={1999},
   month=sep, pages={579–602} }

@article{Kundu2014GeometricSN,
  title={{Geometric Skew Normal Distribution}},
  author={Debasis Kundu},
  journal={Sankhya B},
  year={2014},
  volume={76},
  pages={167 - 189},
  url={https://api.semanticscholar.org/CorpusID:41177050}
}

@article{ElalOlivero2010ALPHASKEWNORMALD,
  title={{ALPHA-SKEW-NORMAL DISTRIBUTION}},
  author={David Elal-Olivero},
  journal={Proyecciones (antofagasta)},
  year={2010},
  volume={29},
  pages={224-240},
  url={https://api.semanticscholar.org/CorpusID:27644304}
}

@article{Mameli2011AGO,
  title={{A Generalization of the Skew-Normal Distribution: The Beta Skew-Normal}},
  author={Valentina Mameli and Monica Musio},
  journal={Communications in Statistics - Theory and Methods},
  year={2011},
  volume={42},
  pages={2229 - 2244},
  url={https://api.semanticscholar.org/CorpusID:88518226}
}

@article{Mudholkar2000TheED,
  title={{The epsilon-skew-normal distribution for analyzing near-normal data}},
  author={Govind S. Mudholkar and Alan David Hutson},
  journal={Journal of Statistical Planning and Inference},
  year={2000},
  volume={83},
  pages={291-309},
  url={https://api.semanticscholar.org/CorpusID:120332325}
}

@article{ArellanoValle2006OnTU,
  title={{On the Unification of Families of Skew‐normal Distributions}},
  author={Reinaldo Boris Arellano-Valle and Adelchi Azzalini},
  journal={Scandinavian Journal of Statistics},
  year={2006},
  volume={33},
  url={https://api.semanticscholar.org/CorpusID:119993511}
}

@article{arnold2000hidden,
  title={{Hidden truncation models}},
  author={Arnold, Barry C and Beaver, Robert J},
  journal={Sankhy{\=a}: The Indian Journal of Statistics, Series A},
  pages={23--35},
  year={2000},
  publisher={JSTOR}
}

@article{vernic2006multivariate,
  title={{Multivariate skew-normal distributions with applications in insurance}},
  author={Vernic, Raluca},
  journal={Insurance: Mathematics and economics},
  volume={38},
  number={2},
  pages={413--426},
  year={2006},
  publisher={Elsevier}
}

@article{carmichael2013asset,
  title={{Asset pricing with skewed-normal return}},
  author={Carmichael, Beno{\i}ˆt and Co{\"e}n, Alain},
  journal={Finance Research Letters},
  volume={10},
  number={2},
  pages={50--57},
  year={2013},
  publisher={Elsevier}
}

@article{fruhwirth2010bayesian,
  title={{Bayesian inference for finite mixtures of univariate and multivariate skew-normal and skew-t distributions}},
  author={Fr{\"u}hwirth-Schnatter, Sylvia and Pyne, Saumyadipta},
  journal={Biostatistics},
  volume={11},
  number={2},
  pages={317--336},
  year={2010},
  publisher={Oxford University Press}
}

@article{rimstad2014skew,
  title={{Skew-Gaussian random fields}},
  author={Rimstad, Kjartan and Omre, Henning},
  journal={Spatial Statistics},
  volume={10},
  pages={43--62},
  year={2014},
  publisher={Elsevier}
}

@article{zhang2010spatial,
  title={{On spatial skew-Gaussian processes and applications}},
  author={Zhang, Hao and El-Shaarawi, Abdel},
  journal={Environmetrics: The official journal of the International Environmetrics Society},
  volume={21},
  number={1},
  pages={33--47},
  year={2010},
  publisher={Wiley Online Library}
}

@inproceedings{DBLP:conf/cvpr/ZhouLSWS024,
  author       = {Xiaoyu Zhou and
                  Zhiwei Lin and
                  Xiaojun Shan and
                  Yongtao Wang and
                  Deqing Sun and
                  Ming{-}Hsuan Yang},
  title        = {{{DrivingGaussian}: Composite {Gaussian Splatting} for Surrounding Dynamic
                  Autonomous Driving Scenes}},
  booktitle    = {{IEEE/CVF} Conference on Computer Vision and Pattern Recognition,
                  {CVPR} 2024, Seattle, WA, USA, June 16-22, 2024},
  pages        = {21634--21643},
  publisher    = {{IEEE}},
  year         = {2024},
  url          = {https://doi.org/10.1109/CVPR52733.2024.02044},
  doi          = {10.1109/CVPR52733.2024.02044},
  bibsource    = {dblp computer science bibliography, https://dblp.org}
}

@article{DBLP:journals/corr/abs-2409-19039,
  author       = {Mahtab Dahaghin and
                  Myrna C. Silva and
                  Kourosh Riahidehkordi and
                  Matteo Toso and
                  Alessio Del Bue},
  title        = {{Gaussian Heritage: {3D} Digitization of Cultural Heritage with Integrated
                  Object Segmentation}},
  journal      = {CoRR},
  volume       = {abs/2409.19039},
  year         = {2024},
  url          = {https://doi.org/10.48550/arXiv.2409.19039},
  doi          = {10.48550/ARXIV.2409.19039},
  eprinttype   = {arXiv},
  eprint       = {2409.19039},
  bibsource    = {dblp computer science bibliography, https://dblp.org}
}

@inproceedings{DBLP:conf/cvpr/KeethaKJYSRL24,
  author       = {Nikhil Varma Keetha and
                  Jay Karhade and
                  Krishna Murthy Jatavallabhula and
                  Gengshan Yang and
                  Sebastian A. Scherer and
                  Deva Ramanan and
                  Jonathon Luiten},
  title        = {{{SplaTAM}: Splat, Track {\&} Map {3D} Gaussians for Dense {RGB-D}
                  {SLAM}}},
  booktitle    = {{IEEE/CVF} Conference on Computer Vision and Pattern Recognition,
                  {CVPR} 2024, Seattle, WA, USA, June 16-22, 2024},
  pages        = {21357--21366},
  publisher    = {{IEEE}},
  year         = {2024},
  url          = {https://doi.org/10.1109/CVPR52733.2024.02018},
  doi          = {10.1109/CVPR52733.2024.02018},
  bibsource    = {dblp computer science bibliography, https://dblp.org}
}

@inproceedings{zwicker2001ewa,
  title={{Ewa volume splatting}},
  author={Zwicker, Matthias and Pfister, Hanspeter and Van Baar, Jeroen and Gross, Markus},
  booktitle={Proceedings Visualization, 2001. VIS'01.},
  pages={29--538},
  year={2001},
  organization={IEEE}
}

@article{zwicker2002ewa,
  title={{Ewa splatting}},
  author={Zwicker, Matthias and Pfister, Hanspeter and Van Baar, Jeroen and Gross, Markus},
  journal={IEEE Transactions on Visualization \& Computer Graphics},
  volume={8},
  number={03},
  pages={223--238},
  year={2002},
  publisher={IEEE Computer Society}
}

@inproceedings{DBLP:conf/cvpr/ZhangZXLX24,
  author       = {Jiahui Zhang and
                  Fangneng Zhan and
                  Muyu Xu and
                  Shijian Lu and
                  Eric P. Xing},
  title        = {{{FreGS}: {3D} {Gaussian Splatting} with Progressive Frequency Regularization}},
  booktitle    = {{IEEE/CVF} Conference on Computer Vision and Pattern Recognition,
                  {CVPR} 2024, Seattle, WA, USA, June 16-22, 2024},
  pages        = {21424--21433},
  publisher    = {{IEEE}},
  year         = {2024},
  url          = {https://doi.org/10.1109/CVPR52733.2024.02024},
  doi          = {10.1109/CVPR52733.2024.02024},
  bibsource    = {dblp computer science bibliography, https://dblp.org}
}

@inproceedings{DBLP:conf/nips/KheradmandRSSTI24,
  author       = {Shakiba Kheradmand and
                  Daniel Rebain and
                  Gopal Sharma and
                  Weiwei Sun and
                  Yang{-}Che Tseng and
                  Hossam Isack and
                  Abhishek Kar and
                  Andrea Tagliasacchi and
                  Kwang Moo Yi},
  editor       = {Amir Globersons and
                  Lester Mackey and
                  Danielle Belgrave and
                  Angela Fan and
                  Ulrich Paquet and
                  Jakub M. Tomczak and
                  Cheng Zhang},
  title        = {{{3D} {Gaussian Splatting} as {Markov Chain Monte Carlo}}},
  booktitle    = {Advances in Neural Information Processing Systems 38: Annual Conference
                  on Neural Information Processing Systems 2024, NeurIPS 2024, Vancouver,
                  BC, Canada, December 10 - 15, 2024},
  year         = {2024},
  url          = {http://papers.nips.cc/paper\_files/paper/2024/hash/93be245fce00a9bb2333c17ceae4b732-Abstract-Conference.html},
  bibsource    = {dblp computer science bibliography, https://dblp.org}
}

@article{DBLP:journals/tog/KnapitschPZK17,
  author       = {Arno Knapitsch and
                  Jaesik Park and
                  Qian{-}Yi Zhou and
                  Vladlen Koltun},
  title        = {{Tanks and temples: benchmarking large-scale scene reconstruction}},
  journal      = {{ACM} Trans. Graph.},
  volume       = {36},
  number       = {4},
  pages        = {78:1--78:13},
  year         = {2017},
  url          = {https://doi.org/10.1145/3072959.3073599},
  doi          = {10.1145/3072959.3073599},
  bibsource    = {dblp computer science bibliography, https://dblp.org}
}

@article{DBLP:journals/tog/HedmanPPFDB18,
  author       = {Peter Hedman and
                  Julien Philip and
                  True Price and
                  Jan{-}Michael Frahm and
                  George Drettakis and
                  Gabriel J. Brostow},
  title        = {{Deep blending for free-viewpoint image-based rendering}},
  journal      = {{ACM} Trans. Graph.},
  volume       = {37},
  number       = {6},
  pages        = {257},
  year         = {2018},
  url          = {https://doi.org/10.1145/3272127.3275084},
  doi          = {10.1145/3272127.3275084},
  bibsource    = {dblp computer science bibliography, https://dblp.org}
}
